%% file: main.tex
\documentclass{article}

\PassOptionsToPackage{numbers, compress}{natbib}

\usepackage[preprint]{neurips_2024}

\usepackage[utf8]{inputenc} 
\usepackage[T1]{fontenc}    
\usepackage{hyperref}       
\usepackage{url}            
\usepackage{booktabs}       
\usepackage{amsfonts}       
\usepackage{nicefrac}       
\usepackage{microtype}      
\usepackage{xcolor}         

\usepackage{graphicx}
\usepackage{subfigure}
\usepackage{longtable}
\usepackage{multirow}
\usepackage{makecell}
\usepackage{rotating}
\usepackage{amsmath}
\usepackage{amssymb}
\usepackage{mathtools}
\usepackage{amsthm}
\usepackage{dsfont}
\usepackage{chngcntr}
\usepackage[capitalize, noabbrev]{cleveref}
\usepackage{pifont}
\usepackage{enumitem}
\setitemize{noitemsep,topsep=0pt,parsep=0pt,partopsep=0pt}

\newcommand{\cmark}{\text{\ding{51}}}
\newcommand{\xmark}{\text{\ding{55}}}

\newcommand{\RB}[1]{\textcolor{blue}{#1}}
\newcommand{\sig}{\boldsymbol{\sigma}}
\newcommand{\muu}{\boldsymbol{\mu}}

\newcommand{\ie}{\textit{i.e.}}

\newcommand{\wrt}{\textit{w.r.t.}}

\DeclareMathOperator{\sign}{sgn}
\DeclareMathOperator{\softmax}{softmax}
\DeclareMathOperator{\diag}{diag}
\DeclareMathOperator{\Var}{Var}
\DeclareMathOperator{\BN}{BN}
\DeclareMathOperator{\LN}{LN}

\DeclareMathOperator{\RMS}{RMSN}
\DeclareMathOperator{\UN}{UN}
\DeclareMathOperator{\ELB}{ELB}

\theoremstyle{plain}
\newtheorem{theorem}{Theorem}[section]

\newtheorem{lemma}[theorem]{Lemma}
\newtheorem{corollary}[theorem]{Corollary}
\theoremstyle{definition}

\theoremstyle{remark}
\newtheorem{remark}[theorem]{Remark}

\usepackage[textsize=tiny]{todonotes}

\title{UnitNorm: Rethinking Normalization for Transformers in Time Series}

\author{%
  Nan Huang \\
  Department of Computer Science\\
  University of North Carolina at Charlotte\\
  Charlotte, NC 28223\\
  \texttt{nhuang1@charlotte.edu} \\
  \And
  Christian K\"{u}mmerle\\
  Department of Computer Science\\
  University of North Carolina at Charlotte\\
  Charlotte, NC 28223\\
  \texttt{kuemmerle@charlotte.edu} \\
  \And
  Xiang Zhang\\
  Department of Computer Science\\
  University of North Carolina at Charlotte\\
  Charlotte, NC 28223\\
  \texttt{xiang.zhang@charlotte.edu} \\
}

\begin{document}

\maketitle

\begin{abstract}
  Normalization techniques are crucial for enhancing Transformer models' performance and stability in time series analysis tasks, yet traditional methods like batch and layer normalization often lead to issues such as token shift, attention shift, and sparse attention.
  We propose UnitNorm, a novel approach that scales input vectors by their norms and modulates attention patterns, effectively circumventing these challenges.
  Grounded in existing normalization frameworks, UnitNorm's effectiveness is demonstrated across diverse time series analysis tasks, including forecasting, classification, and anomaly detection, via a rigorous evaluation on 6 state-of-the-art models and 10 datasets.
  Notably, UnitNorm shows superior performance, especially in scenarios requiring robust attention mechanisms and contextual comprehension, evidenced by significant improvements by up to a 1.46 decrease in MSE for forecasting, and a 4.89\% increase in accuracy for classification.
  This work not only calls for a reevaluation of normalization strategies in time series Transformers but also sets a new direction for enhancing model performance and stability.
  The source code is available at \url{https://anonymous.4open.science/r/UnitNorm-5B84}.
\end{abstract}

\input{010.introduction}

\input{020.motivation}

\input{030.method}

\input{040.experiment}

\input{050.discussion}

\input{060.limitation}

\input{070.conclusion}

\newpage
\bibliography{ref}
\bibliographystyle{unsrtnat}


\newpage
\appendix

\newcounter{appendix}
\counterwithin{figure}{appendix}
\counterwithin{table}{appendix}
\renewcommand{\thefigure}{\Alph{appendix}\arabic{figure}}
\renewcommand{\thetable}{\Alph{appendix}\arabic{table}}
\addtocounter{appendix}{18}
\stepcounter{appendix}

\input{900.appendix}


\newpage
\clearpage
\input{999.checklist}

\end{document}

%% file: 010.introduction.tex
\section{Introduction}

\begin{figure}[ht]
    \vspace{-0.2in}
    \begin{center}
        \centerline{\includegraphics[width=0.8\linewidth]{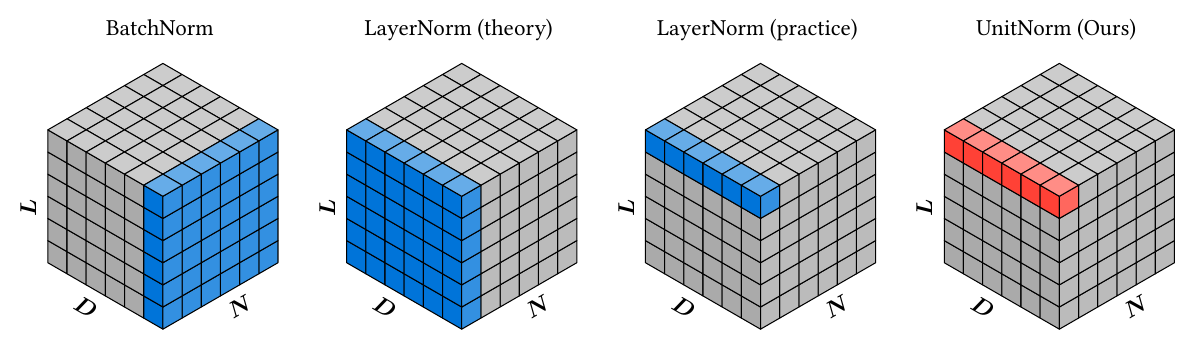}}
        \caption{Scheme of different normalization methods. The input to the normalization layers is batched sequences of token vectors $\mathbf{X} \in \mathbb{R}^{N\times L \times D}$, where $N$ is the batch size, $L$ is the sequence length and $D$ is the dimension of each token vector. The blue sections demonstrate a single slice of the input tensor for computing the mean $\mu$ and variance $\sigma^2$, while the red section shows a single slice of data for computing the vector norm $\left\|\mathbf{x}\right\|$ (see \cref{sec:norm-difference}).}
        \label{fig:norm-comp}
    \end{center}
    \vspace{-0.2in}
\end{figure}

Transformers have revolutionized sequence modeling, demonstrating unparalleled efficacy across diverse fields such as natural language processing (NLP, \citealt{wolf_transformers_2020}), computer vision (CV, \citealt{han_survey_2023}), and recently, time series analysis (TSA, \citealt{wen_transformers_2023}). Central to these models is the representation of data as sequences of token vectors, denoted by $\mathbf{X} \in \mathbb{R}^{N\times L \times D}$, where $N$ stands for batch size, $L$ is the sequence length and $D$ represents the dimensionality of each token.
The core mechanism facilitating the Transformers' ability to model complex dependencies is the attention mechanism. It computes a weighted sum of value vectors $\mathbf{V}$, capturing the sequential relationships between tokens through a scalable dot-product operation of  queries $\mathbf{Q}$ and keys $\mathbf{K}$~\cite{vaswani_attention_2017}:
\begin{equation}
    \label{eq:attention}
    \text{Attention}(\mathbf{Q}, \mathbf{K}, \mathbf{V}) = \softmax\left(\frac{\mathbf{Q}\mathbf{K}^\top}{\sqrt{D}}\right)\mathbf{V},
\end{equation}

To mitigate issues during the training process of Transformers related to vanishing or exploding gradients \cite{lubana_beyond_2021,yang_mean_2017},
Layer Normalization (LayerNorm, $\LN$, \citealt{ba_layer_2016}) plays a significant role and is therefore incorporated at each sub-layer of the architecture (\cref{fig:trans-layer})\footnote{The LayerNorm used in Transformers, referred to as LayerNorm (practice), computes the statistics within each token rather than over the whole batch as LayerNorm (theory) does (\cref{fig:norm-comp}). In this paper, we will refer to the LayerNorm (practice) as LayerNorm if no distinction is made.}. The LayerNorm operation follows the center-and-scale standardization paradigm, by first centering the means to 0 and then rescaling the variances of the input vectors to 1 \cite{ba_layer_2016} such that
\begin{equation}
    \label{eq:layer-norm}
    \LN(\mathbf{X}) = \frac{\mathbf{X} - \boldsymbol{\mu}}{\sqrt{\boldsymbol{\sigma}^2+\varepsilon}},
\end{equation}
where $\boldsymbol{\mu}$ and $\boldsymbol{\sigma}$ are the mean and standard deviation of the input vector $\mathbf{X}$, respectively.

While LayerNorm, compared to other normalization strategies such as batch normalization \cite{ioffe_batch_2015,shen_powernorm_2020,wang_understanding_2022}, has established itself as the dominant normalization strategy in Transformers, dedicated normalization-specific research has mostly focussed on its impact on model convergence \cite{wang_learning_2019}, its inner dynamics \cite{wang_understanding_2022,shen_powernorm_2020} or its location \cite{xiong_layer_2020} within the architecture. On the other hand, only few works touch upon the interaction of normalization with the attention mechanism \cite{kobayashi_incorporating_2021} (see also Related Work \cref{sec:related-work}), which poses specific challenges in TSA (see \cref{sec:challenges}) due to the dot product in attention mechanism.

In this work, we provide a new viewpoint on these challenges by first identifying and formalizing Transformer-specific challenges of normalization techniques, highlighting three key issues. Building on these insights, we introduce a novel normalization technique, UnitNorm, designed to address these challenges effectively.

Our contributions lie in: 1) We originally identify two challenges, namely \textit{token shift} and \textit{attention shift}, and reassess the challenge of \textit{sparse attention} in Transformers \cite{zhai_stabilizing_2023}; 2) We propose a new normalization method, UnitNorm, that can mitigate these issues by design; 3) We empirically validate the effectiveness of UnitNorm on nine datasets spanning three downstream TSA tasks.

%% file: 020.motivation.tex
\section{Challenges in Normalization} \label{sec:challenges}

Transformers rely on attention mechanisms to achieve remarkable performance in time series analysis tasks. However, the interplay between the attention mechanism and the applied normalization methods introduces critical challenges that have yet to be fully addressed. This paper aims to shed light on the complexities of token shift, attention shift, and sparse attention, which arise from the interaction between normalization and the attention mechanism within Transformer models. By presenting a thorough theoretical and empirical analysis, we demonstrate that these challenges are intrinsic to the conventional approaches to normalization, impacting the efficacy of the self-attention mechanism that is central to all Transformer-based architectures.

We explore the relationship between normalization and the attention mechanism by examining a simplified equivalent attention process, where the normalization layer precedes the attention computation (\citealt{zhang_set_2022}, \cref{fig:trans-layer}). This perspective allows for a detailed exploration of how normalization influences the attention scores derived from the query and key vectors. For simplicity, our discussion will center on a singular instance of self-attention within the encoder layer, assuming identical query and key vectors to streamline our analysis (see \cref{sec:feasibility}).

\subsection{Token shift} \label{sec:token-shift}

\begin{table*}[t]
    \centering
    \vspace{-0.2in}
    \caption{Effect of input transformations on the softmax function output. Importance order invariant refers to whether the relative importance of the tokens is preserved. Of all possible input transformations, only the reflection transformation will definitely change the importance order of the tokens.}
    \label{tab:softmax-effect}
    \vspace{0.1in}
    \begin{small}
        {
            \renewcommand{\arraystretch}{1.8}
            \begin{tabular}{llcccr}
                \toprule
                Type       & Function                                                                            & Input                                                                                       & Output                                                                                       & Order invariant?  \\
                \midrule
                None       & $f: x \mapsto x$                                                                    & \raisebox{-0.35\totalheight}{\includegraphics[height=2em]{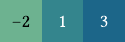}}       & \raisebox{-0.35\totalheight}{\includegraphics[height=2em]{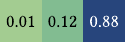}}       &                   \\
                Stretch    & $f: x \mapsto k \cdot x, k \in\ \mathbb{R}^+$                                       & \raisebox{-0.35\totalheight}{\includegraphics[height=2em]{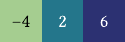}}    & \raisebox{-0.35\totalheight}{\includegraphics[height=2em]{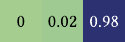}}    & $\cmark$          \\
                Translate  & $f: x \mapsto x + a, a \in \mathbb{R}$                                              & \raisebox{-0.35\totalheight}{\includegraphics[height=2em]{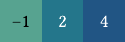}}  & \raisebox{-0.35\totalheight}{\includegraphics[height=2em]{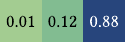}}  & $\cmark$          \\
                Jitter     & $f: x \mapsto x + \varepsilon, \varepsilon \sim \mathcal{N}\left(0,\sigma^2\right)$ & \raisebox{-0.35\totalheight}{\includegraphics[height=2em]{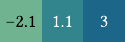}}     & \raisebox{-0.35\totalheight}{\includegraphics[height=2em]{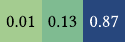}}     & $\cmark / \xmark$ \\
                Reflection & $f: x \mapsto -x$                                                                   & \raisebox{-0.35\totalheight}{\includegraphics[height=2em]{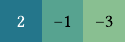}} & \raisebox{-0.35\totalheight}{\includegraphics[height=2em]{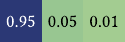}} & $\xmark$          \\
                \bottomrule
            \end{tabular}
        }
    \end{small}
    \vspace{-0.2in}
\end{table*}

\begin{table*}[t]
    \centering
    \caption{Effect of normalization on the attention weight distribution based on empirical results (\cref{fig:word2vec-pair-layer-practice,fig:word2vec-pair-unit}). UnitNorm shows the most faithful representation of the original attention weights that are cross-validated by various metrics as described in \cref{tab:metric-definition}, while center-and-scale normalization significantly alters the attention weights to an extreme extent as depicted in \cref{fig:attention-score-triangle}.}
    \label{tab:attn-dist-shift}
    \vspace{0.1in}
    \begin{small}
        {
            \renewcommand{\arraystretch}{1.8}
            \begin{tabular}{ccccc}
                \toprule
                Normalization    & Chebyshev distance $\downarrow$ & Cosine similarity $\uparrow$ & KL divergence $\downarrow$ & Entropy $\uparrow$ \\
                \midrule
                None (original)  & /                               & /                            & /                          & High               \\
                Center-and-scale & High                            & Low                          & High                       & Very Low           \\
                UnitNorm         & Low                             & High                         & Low                        & High               \\
                \bottomrule
            \end{tabular}
        }
    \end{small}
    \vspace{-0.2in}
\end{table*}

\begin{figure}[htbp]
    \vspace{-0.2in}
    \begin{center}\centerline{\includegraphics[width=0.7\linewidth]{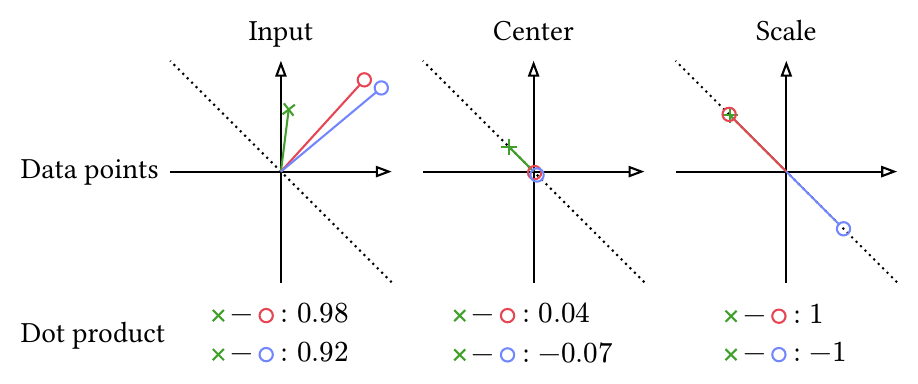}}
        \caption{Case of token shift in LayerNorm. The green cross denotes a query vector, the red and blue circles denote two key vectors. The token shift happens at the centering step of normalization and causes sign flip in dot product, while the scale step will not have such an effect.}
        \label{fig:token-shift}
    \end{center}
    \vspace{-0.2in}
\end{figure}

Previous study \cite{brody_expressivity_2023} has attributed LayerNorm's efficacy to its center-and-scale operations: centering projects the input vectors to a hyperplane orthogonal to $\mathds{1}$ vector, and scaling normalizes the vectors to a unit sphere to prevent any token vector being contained in the convex hull of the others. However, this can significantly alter the orientation of input vectors, especially for those that are near parallel to the hyperplane's norm vector $\mathds{1}$. This alteration impacts the dot product between vectors, potentially leading to sign flips (\cref{fig:token-shift}). Such flips can severely disrupt the softmax function's output (\cref{tab:softmax-effect}), altering the relative importance of tokens \textbf{in a catastrophic way that might convert a significant token into an insignificant one, or vice versa} (\cref{tab:attn-dist-shift}). This issue of significant deviations in attention weight distributions caused by token shift will be further explored in \cref{sec:attention-shift}.

Unfortunately, the propensity for "center-and-scale" normalization to induce such undesirable sign flips in the dot product of vectors is not merely theoretical; it occurs with a high probability, as elucidated by the following theorem.

\begin{theorem}[High probability of sign flip due to center operation]
    \label{thm:dot-product-sign-flip}
    Assume that $\mathbf{x} \sim \mathcal{N}(\muu_{x}, \diag\left(\sig_x^2\right))$, $\mathbf{y} \sim \mathcal{N}(\muu_{y}, \diag\left(\sig_y^2\right))$ are two independent token vectors, with $\muu_{x}, \muu_{y}, \sig_x, \sig_y \in \mathbb{R}^{D}$. Let $\tilde{\mathbf{x}} = \frac{\mathbf{x}-\muu_{x}}{\sig_x}$ and $\tilde{\mathbf{y}} = \frac{\mathbf{y}-\muu_{y}}{\sig_y}$ be the normalized vectors.
    If
    \begin{equation} \label{eq:signflip:theorem:assumption}
        \begin{split}
             & |\muu_{x}^{\top} \muu_{y}| \geq
            12 \left(\sqrt{ \sig_x^{2\top} \sig_y^2} +  \| \sig_x \circ \sig_y \|_{\infty} \right)+ \\
             & 5  \left( \sqrt{\sig_y^{2\top}\! \muu_x^2} \!+\! \sqrt{\sig_x^{2\top}\! \muu_y^2}
            \!+\! \| \sig_y\! \circ\! |\muu_x| \|_{\infty} \!+\! \| \sig_x\! \circ\!  |\muu_y| \|_{\infty}   \right)
        \end{split}
    \end{equation}
    then the probability that the signs of $\mathbf{x}^{\top} \mathbf{y}$ and $\tilde{\mathbf{x}}^{\top} \tilde{\mathbf{y}}$ do not coincide is at least $40 \%$, i.e.,
    \begin{equation}\label{eq:pr_sign}        \Pr(\sign\left(\mathbf{x}^\top\mathbf{y}\right)\ne\sign\left(\tilde{\mathbf{x}}^\top\tilde{\mathbf{y}}\right)) \geq 0.40.
    \end{equation}
\end{theorem}

\begin{remark}
    Derived from the computational methodologies for the statistics of vectors $\mathbf{x}$ and $\mathbf{y}$ (\cref{sec:norm-difference}), BatchNorm posits that the mean vectors are the same so that $\muu_x=\muu_y=\muu$, and similarly $\sig_x^2=\sig_y^2=\sig^2$, while LayerNorm assumes that the mean and standard deviation are shared across feature dimension: $\muu_x = \mu_x\mathds{1}, \muu_y = \mu_y\mathds{1}$ and $\sig_x^2 = \sigma_x^2\mathds{1}, \sig_y^2 = \sigma_y^2\mathds{1}$. Given these assumptions, the condition \eqref{eq:signflip:theorem:assumption} outlined in \cref{thm:dot-product-sign-flip} is satisfied for many token vector distributions. In fact, we show that in the setup of LayerNorm, the condition \eqref{eq:signflip:theorem:assumption} allows for the quotients of token means and standard deviations, i.e., for $\mu_x/\sigma_x$ and $\mu_y/\sigma_y$, to decay as $\Omega( D^{-1/4})$ while still implying a high sign flip probability, cf. \cref{sec:appendix:dim:dep}.
\end{remark}

\cref{thm:dot-product-sign-flip} (see \cref{sec:proofs} for proof) underscores the vulnerability of the "center-and-scale" normalization approach to inadvertently altering the attention mechanism's functionality. The potential for such sign flips, demonstrated with significant likelihood, poses a serious risk to the integrity of the attention scores, as it can lead to a complete reordering of the tokens' importance. We shall see that substantial presence of this issue is not only theoretical, but also empirically validated in the following section.

\subsection{Attention shift} \label{sec:attention-shift}

\begin{figure}[htb]
    \vspace{-0.2in}
    \centering
    \subfigure[Distribution of Chebyshev distance. UnitNorm and RMSNorm preserves the distribution of attention scores, while others significantly alter the distribution.]{\includegraphics[width=0.48\linewidth]{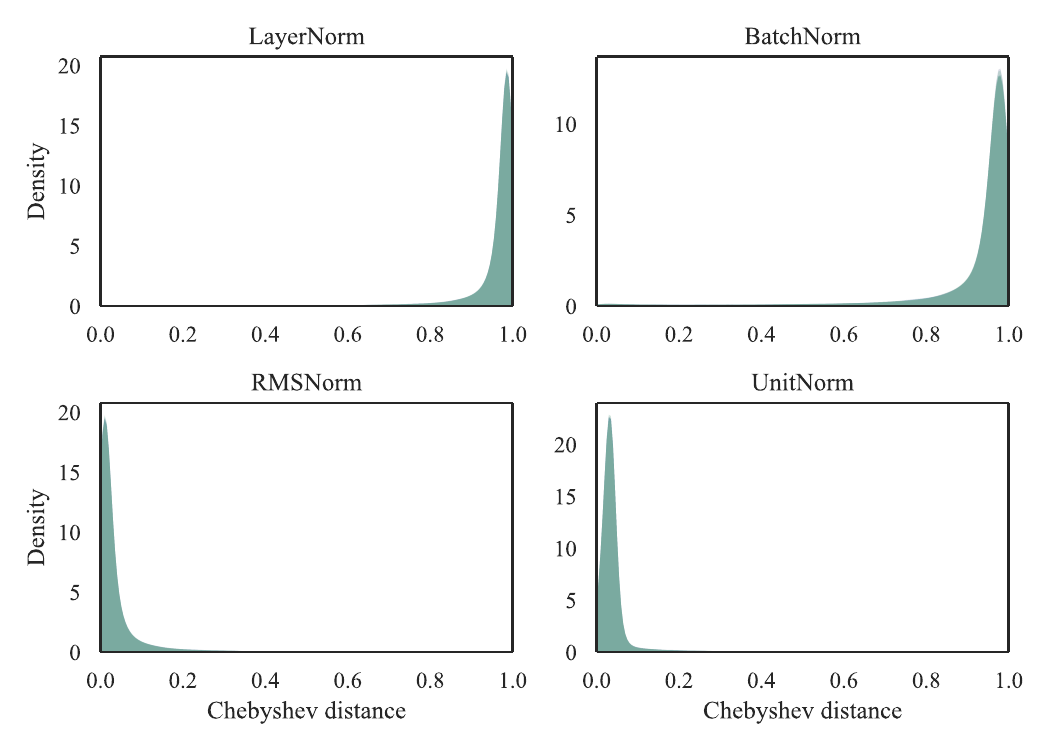} \label{fig:attn-cheb-dist}}
    \hfill 
    \subfigure[Distribution of entropy. UnitNorm and RMSNorm preserves the high entropy of attention scores, while others result in a heavily unbalanced distribution.]{\includegraphics[width=0.48\linewidth]{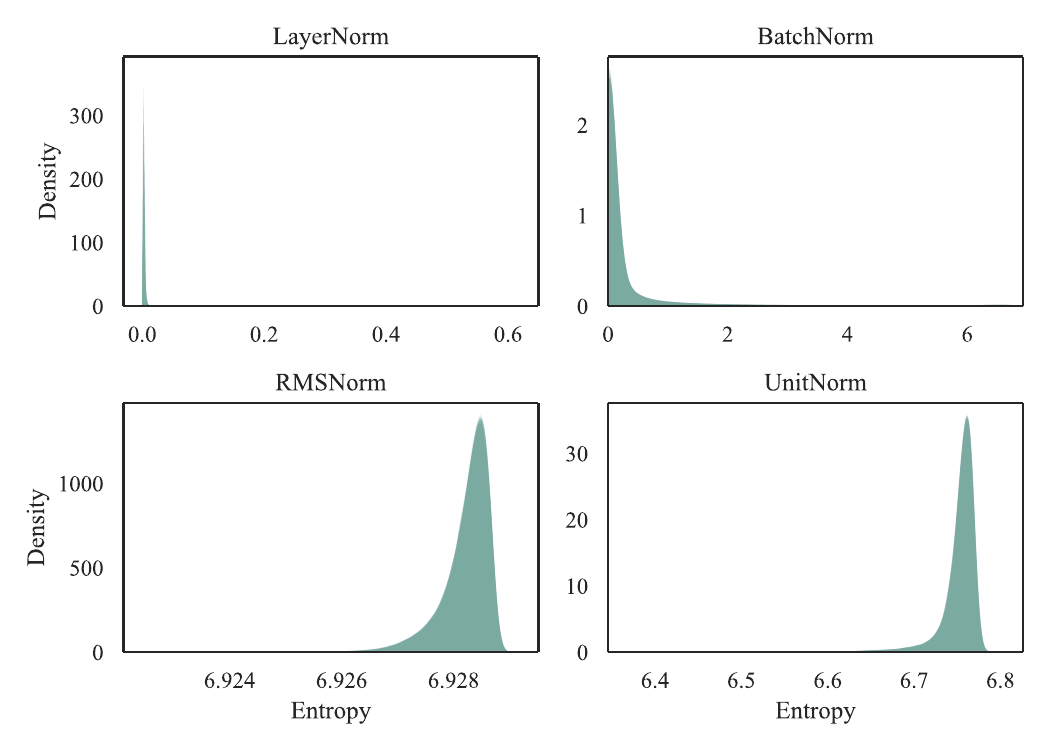} \label{fig:attn-entropy-dist}}
    \caption{Empirical statistics for attention scores after each normalization method. Results from 10 independent experiments are overlaid. $k=1.5$ is used for UnitNorm.}
\end{figure}

Attention shift represents a critical challenge in Transformer models, directly stemming from the token shift issue. This shift perturbs the relative significance of tokens, leading to discrepancies in the attention weights generated from normalized inputs compared to those from the original, un-normalized inputs. To validate the prevalence of attention shift across normalization techniques, we conduct a study utilizing pre-trained Word2Vec embeddings \cite{fares_word_2017}. Our analysis includes a comparison of batch normalization (BatchNorm, $\BN$, \citealt{ioffe_batch_2015}), layer normalization (LayerNorm, $\LN$, \citealt{ba_layer_2016,vaswani_attention_2017}), root mean square layer normalization (RMSNorm, $\RMS$, \citealt{zhang_root_2019}), and our proposed unit normalization (UnitNorm, $\UN$; see \cref{sec:method}).

Our investigation utilizes sequences of token vectors, $\mathbf{X} \in \mathbb{R}^{N \times L \times D}$, as inputs to the normalization layer, where $N$ is the batch size, $L$ is the sequence length, and $D$ is the dimensionality of each token. The attention scores $\mathbf{A} \in \mathbb{R}^{N \times L \times L}$, given as \cref{eq:atten-weight}, are computed for 10 independent sets of 32 batches, each containing 1,024 randomly sampled embeddings from a total of 2 million. The primary goal is to assess the impact of normalization on the fidelity of attention scores $\mathbf{A}$ and $\tilde{\mathbf{A}}$, pre- and post-normalization, using the Chebyshev distance as a metric (\cref{tab:metric-definition}).

\begin{equation}
    \label{eq:atten-weight}
    \mathbf{A}_{n,i} = \softmax\left(\frac{\mathbf{X}_{n,i}\mathbf{X}_{n}^{T}}{\sqrt{D}}\right)
\end{equation}
where $\mathbf{A}_{n,i}\in\mathbb{R}^L$ is the attention scores for the $i$-th anchor token $\mathbf{X}_{n,i}$ to the context sequence $\mathbf{X}_n$ from the $n$-th batch; $\tilde{\mathbf{A}}$ is computed similarly from normalization output $\tilde{\mathbf{X}}$.

The probability density distributions of Chebyshev distances for each normalization method, depicted in \cref{fig:attn-cheb-dist}, reveal significant findings of the inability of maintaining faithful attention distribution of current normalization methods. For BatchNorm and LayerNorm, the Chebyshev distances predominantly span towards the maximum possible value of 1, suggesting a profound alteration in attention weights distribution by normalization. Conversely, UnitNorm and RMSNorm demonstrates a distribution concentrated around zero, indicating minimal disruption to the original attention scores.

The empirical evidence underscores a fundamental issue with current normalization practices in Transformers: they compromise the fidelity of attention scores, leading to distorted relational dynamics between tokens. This distortion challenges not only the model's interpretability but also its ability to learn and adopt complex dependencies accurately.

\subsection{Sparse attention}

The challenge of sparse attention further complicates the normalization landscape in Transformer models. Traditional "center-and-scale" normalization methods often lead to an undesirable concentration of attention scores, effectively pushing the distribution towards one-hot. This is due to fact that centering removes a degree of freedom from the vectors, and only query that are tightly around the $\mathds{1}$ vector can produce uniform attention scores \cite{brody_expressivity_2023}. This can be depicted by the entropy of the attention scores $\mathbf{A}_i$:
\begin{equation}
    \label{eq:entropy}
    H(\mathbf{A}_i) = -\sum_{j=1}^{L}\mathbf{A}_{i,j}\log\mathbf{A}_{i,j}
\end{equation}
A higher entropy value suggests a more uniform attention distribution, enabling models capturing periodicity in time series. Conversely, lower entropy, or a trend towards one-hot distributions, limits its attention to narrow ranges of tokens. While some studies \cite{hyeon-woo_scratching_2022,zhai_stabilizing_2023} in other fields have shown that Transformer models may benefit from capturing longer-range, denser connections, we will show later that such sparse attention is particularly problematic in TSA tasks and requires finer control over the attention patterns.

Analysis of normalization methods through the lens of attention score entropy (\cref{fig:attn-entropy-dist}) reveals a stark contrast in their effects on model behavior. BatchNorm and LayerNorm significantly skew attention distributions towards minimal entropy. This condition not only narrows the model's focus but may also precipitate training instability \cite{zhai_stabilizing_2023}. In contrast, UnitNorm and RMSNorm maintain higher entropy levels, suggesting a more balanced and contextually aware attention mechanism. Notably, the key deviation in attention entropy between UnitNorm and RMSNorm is the former's ability to modulate the entropy pattern by adjusting the $k$ parameter, as discussed in \cref{sec:method}, while RMSNorm maintains a consistent high entropy level close to the theoretical upper bound $\log {L}$.

%% file: 030.method.tex
\section{Methodology} \label{sec:method}

To mitigate the challenges identified with traditional normalization methods, we introduce a novel approach called \textbf{unit normalization (UnitNorm, $\UN$)}, formulated such that
\begin{equation}
    \label{eq:unit-norm}
    \UN(\mathbf{X}) = D^{\frac{k}{2}}\frac{\mathbf{X}}{\left\|\mathbf{X}\right\|_2}.
\end{equation}
Diverging from the conventional center-and-scale paradigm, UnitNorm omits the center operation entirely. Similar to RMSNorm, UnitNorm focuses solely on scaling the input vectors, first normalizing the input vectors by their $\ell^2$ norm. However, UnitNorm is different to RMSNorm through subsequently scaling them by a factor of $D^{\frac{k}{2}}$, where $k$ is a hyperparameter dictating the sparsity of the resulting attention scores.

\subsection{Theoretical foundation}

UnitNorm is theoretically grounded as a variant of LayerNorm and RMSNorm. Specifically, when taking $k=1$, UnitNorm is effectively acting as LayerNorm with asserted zero mean, and the RMSNorm can be seen as a special case of UnitNorm with $k=1$.

This equivalence suggests that UnitNorm inherits the beneficial properties of LayerNorm and RMSNorm, such as mitigating gradient vanishing or exploding and stabilizing training. It maintains consistent forward pass and gradient propagation regardless of scaling in learnable parameters, while scaling down the gradient to these parameters when they are large (proved in \cref{sec:proofs}), thus ensuring stable training conditions:
\begin{theorem}[UnitNorm preseves the gradient to the input and stablize the gradient to the learnable parameters]
    \label{thm:unit-norm-gradient}
    Given the output of an affine transformation $\mathbf{x} = \mathbf{W}\mathbf{v} + \mathbf{b}$, where $\mathbf{W}$ and $\mathbf{b}$ are learnable parameters. If $\mathbf{x}' = (\alpha\mathbf{W})\mathbf{v} + (\alpha\mathbf{b})$, then the output of UnitNorm is unchanged, \ie, $\tilde{\mathbf{x}}' = \tilde{\mathbf{x}}$, while the gradients to loss $\mathcal{L}$ are given as follows:
    \begin{equation}
        \begin{aligned}
            \frac{\partial \mathcal{L}}{\partial \tilde{\mathbf{x}}'} \cdot \frac{\partial \tilde{\mathbf{x}}'}{\partial (\alpha\mathbf{W})}
             & = \frac{1}{\alpha} \cdot \frac{\partial \mathcal{L}}{\partial \tilde{\mathbf{x}}} \cdot \frac{\partial \tilde{\mathbf{x}}}{\partial \mathbf{W}}
             &                                                                                                                                                 & = \frac{1}{\alpha} \cdot \frac{\partial \mathcal{L}}{\partial \tilde{\mathbf{x}}} \cdot \mathbf{J} \mathbf{v}^\top \\
            \frac{\partial \mathcal{L}}{\partial \tilde{\mathbf{x}}'} \cdot \frac{\partial \tilde{\mathbf{x}}'}{\partial (\alpha\mathbf{b})}
             & = \frac{1}{\alpha} \cdot \frac{\partial \mathcal{L}}{\partial \tilde{\mathbf{x}}} \cdot \frac{\partial \tilde{\mathbf{x}}}{\partial \mathbf{b}}
             &                                                                                                                                                 & = \frac{1}{\alpha} \cdot \frac{\partial \mathcal{L}}{\partial \tilde{\mathbf{x}}} \cdot \mathbf{J}                 \\
            \frac{\partial \mathcal{L}}{\partial \tilde{\mathbf{x}}'} \cdot \frac{\partial \tilde{\mathbf{x}}'}{\partial \mathbf{v}}
             & = \frac{\partial \mathcal{L}}{\partial \tilde{\mathbf{x}}} \cdot \frac{\partial \tilde{\mathbf{x}}}{\partial \mathbf{v}}
             &                                                                                                                                                 & = \frac{\partial \mathcal{L}}{\partial \tilde{\mathbf{x}}} \cdot \mathbf{J} \mathbf{W}^\top
        \end{aligned}
    \end{equation}
    where $\mathbf{J}$ is the Jacobian matrix of $\tilde{\mathbf{x}}$ \wrt $\mathbf{x}$.
\end{theorem}

\subsection{Overcoming defects}

While UnitNorm shares similar learning dynamics with RMSNorm, by omitting the center operation, it preserves the directions of original input vectors, directly addressing the token and attention shift problems by maintaining the dot product's sign (\cref{fig:demo}). This allows UnitNorm to serve as a drop-in replacement for LayerNorm and RMSNorm in time series Transformer architectures, requiring no structural modifications.

\begin{figure}[tbp]
    \vspace{-0.2in}
    \centering
    \includegraphics[width=0.6\linewidth]{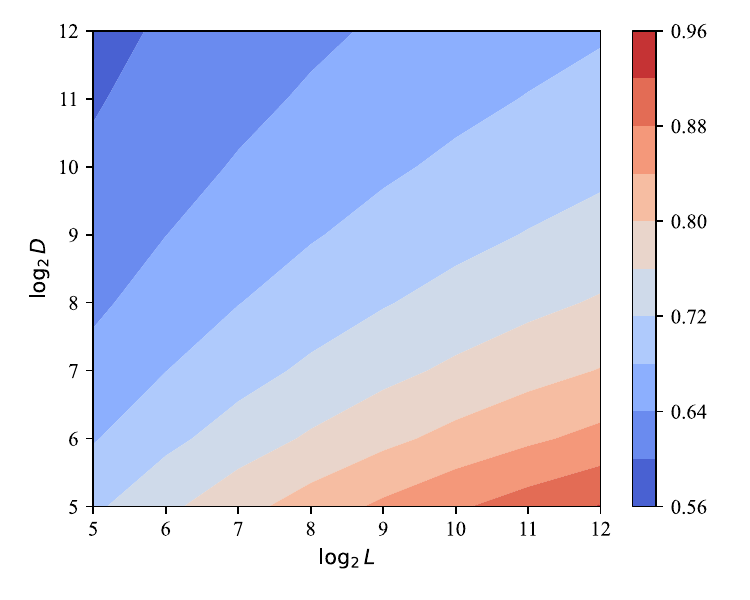}
    \vspace{-0.2in}
    \caption{Landscape of $k_{50}$ for different $L, D$. The $k_{50}$ is the value of $k$ that achieves an ELB of half of the theoretical maximum $\log{L}$ for a given $L, D$ pair. The landscape of $k_{50}$ is rather smooth and insensitive to the sequence length $L$, indicating UnitNorm with fixed $k$ can be applied to sequences with variable length without significant change in the attention pattern.}
    \label{fig:ELB-k-50}
    \vspace{-0.1in}
\end{figure}

Additionally, UnitNorm confronts the sparse attention issue by introducing an entropy lower bound (ELB) for attention scores, modulated by the hyperparameter $k$ (proved in \cref{sec:proofs}). This feature enables the control of attention patterns, from dense as uniform to sparse as one-hot, offering versatility in modeling attention dynamics:
\begin{theorem}[UnitNorm guarantees an entropy lower bound independent of the input]
    \label{thm:unit-norm-entropy}
    For a given set of $L,D$ and a given $k$, there exists an entropy lower bound (ELB) of the attention scores, \ie
    \begin{equation}
        \label{eq:unit-norm-entropy}
        \ELB(k;L,D) = \log\left(L-1+e^d\right) - \frac{de^d}{L-1+e^d},
    \end{equation}
    where $d = 2D^{k-\frac{1}{2}}$.
\end{theorem}

\begin{corollary}[The ELB of UnitNorm can be any possible value by modulating $k$]
    \label{lem:unit-norm-entropy-limit}
    The ELB is a monotonically decreasing function of $k$ for a given $L,D$. Furthermore, it is bounded that $\forall k$:
    \begin{equation}
        0 < \ELB(k;L,D) < \log{L}
    \end{equation}
\end{corollary}

The adaptability of UnitNorm is further exemplified by its applicability across variable sequence lengths, with the entropy lower bound's sensitivity to $k$ remaining relatively consistent irrespective of sequence length (\cref{fig:ELB-LD-curve}), along with the smooth landscape of $k_{50}$, the value of $k$ that achieves an ELB of $\frac{1}{2}\log{L}$ for a given $L, D$ pair (\cref{fig:ELB-k-50}), particularly with larger $D$. This property, combined with the option of setting $k$ as a learnable parameter, empowers the model to dynamically adjust its attention pattern, optimizing performance across different tasks and data sets.

%% file: 040.experiment.tex
\section{Experiments} \label{sec:experiments}

In our experimental evaluation, UnitNorm is rigorously tested across a spectrum of TSA tasks to illustrate its theoretical advantages in practical applications, including long term forecasting (ETTh1, ETTh2, ECL, Exchange), classification (FaceDetection, Heartbeat, PEMS-SF, UWaveGestureLibrary) and anomaly detection (MSL). We integrate UnitNorm into various Transformer models, namely Crossformer \cite{zhang_crossformer_2022}, FEDformer \cite{zhou_fedformer_2022}, Informer \cite{zhou_informer_2021}, PatchTST \cite{nie_time_2022} and the vanilla Transformer \cite{vaswani_attention_2017}, all with same set of hyperparameter as described in \cite{wu_timesnet_2023}. For comparison, we also include BatchNorm, LayerNorm, RMSNorm and various settings of UnitNorm (see figure legends). By doing so, we aim to demonstrate its superior ability to address normalization-related challenges, enhancing model performance in these tasks. Detailed experimental settings and full results are provided in \cref{tab:long-term-forecast-dataset,tab:classification-dataset,tab:anomaly-detection-dataset,tab:long-term-forecast,tab:classification,tab:anomaly-detection}. Below, we outline the significance of these tasks and the specific benefits UnitNorm brings.

\begin{figure}[htb]
    \centering
    \includegraphics[width=\linewidth]{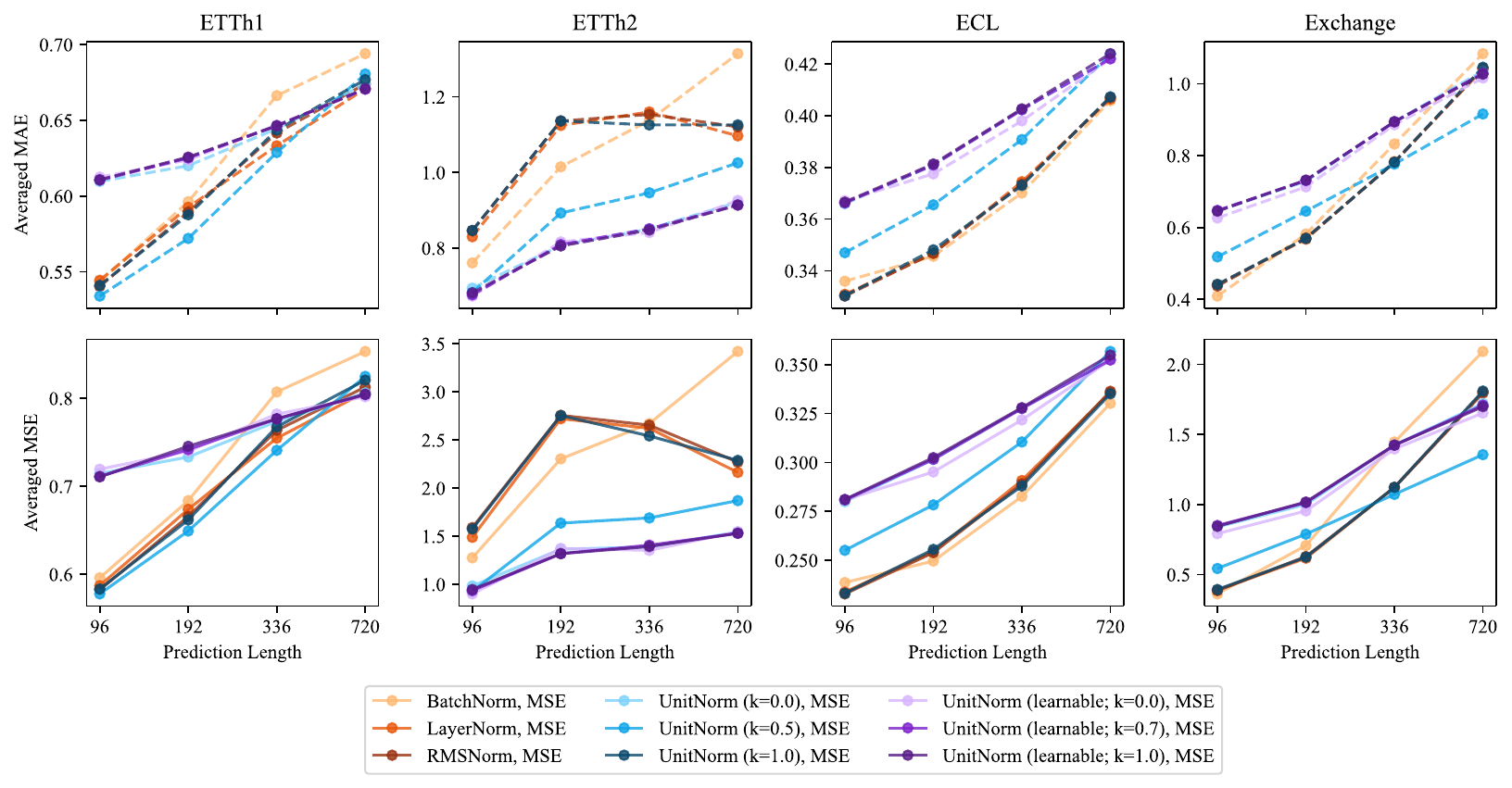}
    \vspace{-0.2in}
    \caption{Average rank of normalization methods on the long-term forecasting tasks. X-axis: number of tokens to forecast, Y-axis: average rank over models. Ranks are computed based on the MAE or MSE of each model on each task with different normalization methods (lower is better). UnitNorm and UnitNorm (learnable) achieve better results with the increase of prediction horizon, and have a slower increase in prediction error compared to other normalization methods.}
    \label{fig:long-term-forecast}
\end{figure}

\textbf{Long-term forecasting}: Long-term forecasting represents a significant challenge for Transformer models, primarily due to the difficulty in maintaining periodic pattern recognition over extended sequences \cite{li_towards_2023}. The conventional normalization methods often exacerbate the sparse attention problem, hindering the model's capability to capture periodicity. In contrast, UnitNorm demonstrates exceptional performance in mitigating this issue, as indicated by its superior rank over longer prediction horizon and slower increase in prediction error across various datasets (\cref{fig:long-term-forecast}). With a maximum increase of 1.46/0.45 in MSE/MAE on ETTh2, and 1.27/0.36 in MSE/MAE on Exchange at the longest prediction horizon, it substantiates UnitNorm's ability to preserve the attention mechanism's effectiveness, even with increasing prediction horizons, due to its ability to maintain a balanced attention distribution and omission of token shift and attention shift disturbances.

\textbf{Classification}: In classification tasks, the key challenge lies in effectively capturing long-range dependencies within sequences \cite{vyas_transdbc_2022}, a task at which Transformers excel. However, the efficacy of this capability can be significantly impacted by the choice of normalization method. UnitNorm, with its unique approach to normalization, has been shown to enhance model performance across multiple datasets, outperforming traditional methods in 3 out of 4 datasets on average (\cref{fig:classification}), with a significant increase in accuracy of up to 4.90\% on UWaveGestureLibrary, 1.95\% on Heartbeat and 0.48\% on FaceDetection. This underscores the versatility of UnitNorm in adapting to varied datasets, offering improved accuracy by enabling a more robust, contextually aware attention mechanism.

\textbf{Anomaly detection}: Anomaly detection in time series data demands robust model sensitivity to subtle deviations \cite{haq_transnas-tsad_2023,yang_ddmt_2023}, a requirement often compromised by normalization-induced shifts in attention. The token and attention shift problems, in particular, pose significant challenges in learning stable representations. UnitNorm addresses these challenges head-on, providing a more stable foundation for anomaly detection models to operate on, therefore gaining a maximum of 7.32\% in recall, 5.58\% in F-score, and 2.81\% in precision. Its effectiveness is dominant in all accuracy, recall, precision, and F-score metrics (\cref{fig:anomaly-detection}), showcasing its capacity to facilitate more accurate and reliable time series modeling for anomaly detection.

%% file: 050.discussion.tex
\section{Discussion}\label{sec:discussion}

This study introduces UnitNorm, a normalization method tailored to Transformers that addresses the inherent limitations of currently prevalent normalization techniques. Here, we reflect on the broader implications of our findings and chart potential avenues for future research.

\subsection{Related work} \label{sec:related-work}

The development of effective normalization techniques is crucial in the optimization of neural network training, particularly for Transformer models \cite{wang_learning_2019}. This section reviews notable contributions in this field, providing a context for our proposed UnitNorm method.

The quest for effective normalization in neural networks, particularly Transformers, is ongoing, with significant strides made in understanding and optimizing these models' training dynamics \cite{wang_learning_2019}. However, we can see that UnitNorm is fundamentally different from existing research directions, and provides a novel perspective on the role of normalization in Transformer models.

\textbf{Normalization layer placement}: The discourse around normalization in Transformer models has predominantly revolved around its placement: Post-Layer Normalization (Post-LN) versus Pre-Layer Normalization (Pre-LN), highlighting its impact on training stability and gradient flow \cite{xiong_layer_2020}. Our approach with UnitNorm shifts focus from placement to the essence of normalization itself, emphasizing the importance of preserving token vector direction being crucial for the attention mechanism, a perspective that can be applied to both Post-LN and Pre-LN Transformers.

\textbf{Normalization layer design for Transformers}: Following RMSNorm \cite{zhang_root_2019}, UnitNorm eliminates the center operation in normalization and alleviate token shift and attention shift problem, while owning a fundamental departure by introducing a hyperparameter $k$ to modulate the sparsity of attention scores. This design choice is unique to aid capturing periodicity in time series data.

\textbf{Normalization on model weight}: Another parallel can be drawn with Weight Normalization \cite{salimans_weight_2016}, which, despite its computational similarity to UnitNorm, applies to model parameters rather than inputs. Weight Normalization also focused on re-parameterizing for training acceleration, and thus still requires a modified BatchNorm for normalization on layer data. This distinction underscores UnitNorm's unique approach to addressing the input-specific challenges in Transformers, diverging from methods that primarily focused on parameter optimization.

\subsection{Adopting UnitNorm in Transformer models}

UnitNorm invites reconsideration of standard normalization practices in Transformers, suggesting alternatives that might enhance model performance and stability. Its simplicity and versatility suggest it could be readily adopted across various Transformer applications. The broader impact of UnitNorm lies in its potential to improve the applicability and efficiency of Transformers in fields where precision and model stability are paramount. By addressing specific normalization-related challenges, UnitNorm can make Transformers more suitable for tasks with complex sequential relationships.

%% file: 060.limitation.tex
\section{Limitations} \label{sec:limitations}

While UnitNorm represents a significant advancement in normalization techniques for Transformers, several areas still warrant further investigation:
\begin{itemize}[leftmargin=*]
    \item \textbf{Dynamic and Adaptive Normalization}: Investigating UnitNorm's adaptability, particularly the dynamic adjustment of the hyperparameter $k$, could lead to performance optimizations tailored to specific tasks.
    \item \textbf{Broader Application Scope}: Extending the application of UnitNorm beyond Transformers to other neural network architectures could provide valuable insights into the fundamental principles of normalization across deep learning models.
    \item \textbf{Cross Domain Validation}: Applying UnitNorm across diverse domains and challenging datasets beyond TSA, e.g., NLP \cite{brown_language_2020,devlin_bert_2019} and CV \cite{dosovitskiy_image_2020}, will further elucidate its effectiveness and generalizability, providing insights into its broad utility in deep learning.
    \item \textbf{Problem characterization}: Understanding how and what certain dataset characteristics influence the efficacy of normalization methods, including quantitatively assess the presence of token shift, attention shift, and sparse attention in the dynamic interplay of attention mechanisms and normalization during training, can guide the community in selecting appropriate techniques for varied deep learning challenges.
\end{itemize}

Much as UnitNorm marks a promising advancement in normalization for Transformers, its exploration is far from complete. The limitations identified herein not only highlight the need for further empirical validation across domains but also the potential for refining and extending the methodology to accommodate a wider array of neural network architectures and applications.

%% file: 070.conclusion.tex
\section{Conclusion} \label{sec:conclusion}

Through the introduction of UnitNorm, this study challenges prevailing norms around normalization in Transformer models for TSA tasks, underscoring the importance of a tailored approach to normalization. UnitNorm's innovative strategy, eschewing the conventional center operation, directly addresses the critical issues of token shift, attention shift that we have identified, along with sparse attention, which have been overlooked in traditional normalization practices.

Our contribution extends beyond the theoretical introduction of UnitNorm; it includes empirical evidence showcasing its efficacy across various tasks, setting a new precedent for normalization techniques within the Transformer architecture. By facilitating a more stable and faithful representation learning, UnitNorm paves the way for enhanced performance and broader applicability of Transformer models in complex sequential data analysis.

While there are also many potential ethical consequences of our work, given the theoretical nature of this work, a detailed discussion on ethical impacts falls beyond its scope. Future endeavors that leverage UnitNorm in application-specific contexts should carefully assess these considerations.

%% file: 900.appendix.tex
\input{910.appendix.sign-flip}

\input{920.appendix.proof}

\input{930.appendix.discussion}

\input{940.appendix.figure}

\input{950.appendix.table}

%% file: 910.appendix.sign-flip.tex
\section{Dimension Dependence of Sign-Flip Probability} \label{sec:appendix:dim:dep}
We recall that \cref{thm:dot-product-sign-flip} provided a condition for token vector means and variances, condition \eqref{eq:signflip:theorem:assumption}, to imply that the sign of the token dot product $\mathbf{x}^{\top} \mathbf{y}$ is flipped by center-and-scale standardization as in LayerNorm \cite{ba_layer_2016}.

In this section, we elucidate the dimension dependence of the required relationship between token means and standard deviations implied by this condition in the case of shared means and standard deviations across feature dimensions, such as implicitly assumed by LayerNorm.
\begin{corollary} \label{cor:LayerNorm}
    Assume that the mean and variance vectors of independent token vectors $\mathbf{x}$ and $\mathbf{y}$ satisfy $\muu_x = \mu_x\mathds{1}, \muu_y = \mu_y\mathds{1}$ and $\sig_x^2 = \sigma_x^2\mathds{1}, \sig_y^2 = \sigma_y^2\mathds{1}$. Then the mean-variance condition \eqref{eq:signflip:theorem:assumption} of \cref{thm:dot-product-sign-flip} is satisfied for all $L \geq 77$ if
    \begin{equation} \label{eq:assumption:Appendixlemma}
        \frac{\mu_x}{\sigma_x} \geq \frac{6}{D^{1/4}} \quad \text{ and } \quad \frac{\mu_y}{\sigma_y} \geq \frac{6}{D^{1/4}},
    \end{equation}
    Furthermore, if additionally the independent token vectors are distributed as $\mathbf{x} \sim \mathcal{N}(\muu_{x}, \diag\left(\sig_x^2\right))$, $\mathbf{y} \sim \mathcal{N}(\muu_{y}, \diag\left(\sig_y^2\right))$, then the dot product $\tilde{\mathbf{x}}^{\top} \tilde{\mathbf{y}}$ of normalized vectors $\tilde{\mathbf{x}} = \frac{\mathbf{x}-\muu_{x}}{\sig_x}$ and $\tilde{\mathbf{y}} = \frac{\mathbf{y}-\muu_{y}}{\sig_y}$ attains a sign flip with respect to the original inner products $\mathbf{x}^{\top} \mathbf{y}$ with probability of at least $40\%$.
\end{corollary}
\cref{cor:LayerNorm} implies that for high-dimensional token vectors with $D \gg 1$, it might become easier to satisfy \eqref{eq:assumption:Appendixlemma} given an empirical token distribution, which means that sign flips of dot products after LayerNorm-style normalization might become even more prevalent in that case.

\begin{proof}[Proof of \cref{cor:LayerNorm}]
    For the case of $\muu_x = \mu_x\mathds{1}, \muu_y = \mu_y\mathds{1}$ and $\sig_x^2 = \sigma_x^2\mathds{1}, \sig_y^2 = \sigma_y^2\mathds{1}$, it follows that
    \[
        \begin{split}
                 & 12 \left(\sqrt{ \sig_x^{2\top} \sig_y^2} +  \| \sig_x \circ \sig_y \|_{\infty} \right)+ 5  \left( \sqrt{\sig_y^{2\top}\! \muu_x^2} \!+\! \sqrt{\sig_x^{2\top}\! \muu_y^2}
            \!+\! \| \sig_y\! \circ\! |\muu_x| \|_{\infty} \!+\! \| \sig_x\! \circ\!  |\muu_y| \|_{\infty}   \right)                                                                                                                                      \\
            =    & 12 \left( \sqrt{D \sigma_x^2 \sigma_y^2 } + \sigma_x \sigma_y   \right) + 5 \left( \sqrt{ D \sigma_y^2 \mu_x^2  } + \sqrt{ D \sigma_x^2 \mu_y^2  } + \sigma_y |\mu_x| + \sigma_x |\mu_y|  \right)                                      \\
            \leq & 12 \left( \sqrt{D \frac{D \mu_x^2 \mu_y^2}{36^2}} + \frac{D^{1/4} \mu_x}{6}  \frac{D^{1/4} \mu_y}{6}  \right)                                                                                                                          \\
                 & + 5 \left( \sqrt{ D \frac{D^{1/2} \mu_y^2}{36} \mu_x^2  } + \sqrt{ D \frac{D^{1/2} \mu_x^2}{36} \mu_y^2  } +\frac{D^{1/4} \mu_y}{6} |\mu_x| + \frac{D^{1/4} \mu_x}{6} |\mu_y|  \right)                                                 \\
            =    & 12 \left( D  \frac{\mu_x \mu_y}{36} + D^{1/2} \frac{ \mu_x \mu_y}{36}    \right) + \frac{5}{6} \left( \sqrt{ D D^{1/2} \mu_y^2 \mu_x^2  } + \sqrt{ D D^{1/2} \mu_x^2 \mu_y^2  } +D^{1/4} \mu_y|\mu_x| + D^{1/4} \mu_x |\mu_y|  \right) \\
            =    & \mu_x \mu_y \left(\frac{1}{3} D + \frac{1}{3}D^{1/2} + \frac{5}{3}(D^{3/4} + D^{1/4}) \right) \leq D \mu_x \mu_y = |\muu_x^{\top} \muu_y|.
        \end{split}
    \]
    Here, we used in the first inequality the assumption \cref{eq:assumption:Appendixlemma} and the fact that $\frac{1}{3}D^{1/2} + \frac{5}{3}(D^{3/4} + D^{1/4}) \leq \frac{2}{3} D$ for $D \geq 77$ in the last inequality. The last assertion of the theorem then follows by application of \cref{thm:dot-product-sign-flip}.
\end{proof}

%% file: 920.appendix.proof.tex
\section{Proofs} \label{sec:proofs}
In this section, we detail the proofs of the theoretical results of this paper. In particular, we present the proofs of \cref{thm:dot-product-sign-flip}, \cref{thm:unit-norm-gradient}, \cref{thm:unit-norm-entropy}, \cref{lem:unit-norm-entropy-limit}, as well as of auxiliary lemmas.

\subsection{Proof of \cref{thm:dot-product-sign-flip}}

\begin{proof}[Proof of \cref{thm:dot-product-sign-flip}]
    Let  $\mathbf{x} \sim \mathcal{N}(\muu_{x}, \diag\left(\sig_x^2\right))$ and $\mathbf{y} \sim \mathcal{N}(\muu_{y}, \diag\left(\sig_y^2\right))$ be independent, and write $\mathbf{x}= \begin{pmatrix}
            X_1, & \ldots, X_D
        \end{pmatrix}$
    and $\mathbf{y}= \begin{pmatrix}
            Y_1, & \ldots, Y_D
        \end{pmatrix}$, respectively.

    then we can compute the expectation $\mathbb{E}\left[\mathbf{x}^\top\mathbf{y}\right]$ of the dot product of $\mathbf{x}$ and $\mathbf{y}$ as
    \begin{equation*}
        \begin{aligned}
            \mathbb{E}\left[\mathbf{x}^\top\mathbf{y}\right] = & \mathbb{E}\left[\sum_{i=1}^{D}{X_i Y_i}\right]                       \\
            =                                                  & \sum_{i=1}^{D}{\mathbb{E}\left[X_i Y_i\right]}                       \\
            =                                                  & \sum_{i=1}^{D}{\mathbb{E}\left[X_i\right]\mathbb{E}\left[Y_i\right]} \\
            =                                                  & \sum_{i=1}^{D}{(\boldsymbol{\mu}_x)_i(\boldsymbol{\mu}_y)_i}         \\
            =                                                  & \boldsymbol{\mu}_x^\top\boldsymbol{\mu}_y.
        \end{aligned}
    \end{equation*}

    \begin{equation}
        \label{eq:dot-product-variance-partial}
        \begin{aligned}
            \Var\left(\mathbf{x}^\top\mathbf{y}\right) = & \mathbb{E}\left[\left(\mathbf{x}^\top\mathbf{y}\right)^2\right] - \left(\mathbb{E}\left[\mathbf{x}^\top\mathbf{y}\right]\right)^2           \\
            =                                            & \mathbb{E}\left[\left(\sum_{i=1}^{D}{X_i Y_i}\right)^2\right] - \left(\boldsymbol{\mu}_x^\top\boldsymbol{\mu}_y\right)^2                    \\
            =                                            & \sum_{i,j=1}^{D}{\mathbb{E}\left[X_i Y_i X_j X_j \right]} - \left(\boldsymbol{\mu}_x^\top\boldsymbol{\mu}_y\right)^2                        \\
            =                                            & \sum_{i,j=1}^{D}{\mathbb{E}\left[X_i X_j \right]\mathbb{E}\left[ Y_i Y_j\right]} - \left(\boldsymbol{\mu}_x^\top\boldsymbol{\mu}_y\right)^2
        \end{aligned}
    \end{equation}
    By definition of covariance, we have $\boldsymbol{\sigma}_x = \mathbb{E}\left[\mathbf{x}\mathbf{x}^\top\right] - \boldsymbol{\mu}_x\boldsymbol{\mu}_x^\top$, and here $\boldsymbol{\sigma}_x = \diag(\boldsymbol{\sigma}_x^2)$, then \cref{eq:dot-product-variance-partial} can be simplified as follows:
    \begin{equation}
        \label{eq:dot-product-variance}
        \begin{aligned}
            \Var\left(\mathbf{x}^\top\mathbf{y}\right) = & \sum_{i=1}^{D}{\left(\boldsymbol{\sigma}_x + \boldsymbol{\mu}_x\boldsymbol{\mu}_x^\top\right)_{ij}\left(\boldsymbol{\sigma}_y + \boldsymbol{\mu}_y\boldsymbol{\mu}_y^\top\right)_{ij}} - \left(\boldsymbol{\mu}_x^\top\boldsymbol{\mu}_y\right)^2                                                                                                                                                \\
            =                                            & \left<\boldsymbol{\sigma}_x, \boldsymbol{\sigma}_y\right>_F + \left<\boldsymbol{\mu}_x\boldsymbol{\mu}_x^\top, \boldsymbol{\sigma}_y\right>_F + \left<\boldsymbol{\sigma}_x, \boldsymbol{\mu}_y\boldsymbol{\mu}_y^\top\right>_F + \left<\boldsymbol{\mu}_x\boldsymbol{\mu}_x^\top, \boldsymbol{\mu}_x\boldsymbol{\mu}_x^\top\right>_F - \left(\boldsymbol{\mu}_x^\top\boldsymbol{\mu}_y\right)^2 \\
            =                                            & \left(\boldsymbol{\sigma}_x^2\right)^\top\left(\boldsymbol{\sigma}_y^2\right) + \left(\boldsymbol{\sigma}_y^2\right)^\top\left(\boldsymbol{\mu}_x^2\right) + \left(\boldsymbol{\sigma}_x^2\right)^\top\left(\boldsymbol{\mu}_y^2\right)
        \end{aligned}
    \end{equation}
    where $\left<\cdot,\cdot\right>_F$ is the Frobenius inner product.

    Consider now the normalized random vectors $\tilde{\mathbf{x}} = \frac{\mathbf{x}-\muu_{x}}{\sig_x}$ and $\tilde{\mathbf{y}} = \frac{\mathbf{y}-\muu_{y}}{\sig_y}$. Due to the Gaussianity assumption on $\mathbf{x}$ and $\mathbf{y}$, it follows that the normalized vectors are also Gaussian, and in particular, are distributed as $\tilde{\mathbf{x}},\tilde{\mathbf{y}} \sim \mathcal{N}( \mathbf{0}, \mathbb{I})$. Plugging the respective mean and variance values into the formulas for the expectation and variance for dot products above, we obtain that
    \begin{equation}
        \mathbb{E}\left[\tilde{\mathbf{x}}^\top\tilde{\mathbf{y}}\right] = 0 \quad \text{ and }\quad
        \Var\left(\tilde{\mathbf{x}}^\top\tilde{\mathbf{y}}\right) = 1
    \end{equation}
    As $\tilde{\mathbf{x}}^\top\tilde{\mathbf{y}}$ is a symmetric random variable, it follows that
    \begin{equation} \label{eq:dotproduct:05}
        \Pr\left(\tilde{\mathbf{x}}^\top\tilde{\mathbf{y}}\right) = 0.5.
    \end{equation}

    Next, due to the definition of the random vectors $\mathbf{x}$ and $\mathbf{y}$, it holds that  $\mathbf{x}^\top\mathbf{y} = \sum_{i=1}^D X_i Y_i $, where $X_i \sim \mathcal{N}((\muu_x)_i, (\sig_x)_i^2)$ and $Y_i \sim \mathcal{N}((\muu_y)_i, (\sig_y)_i^2)$ are independent normal random variables. Going forward, we will use the $\psi_1$-Orlicz norm
    \begin{equation} \label{eq:def:psi1}
        \| X\|_{\psi_1} := \inf \{t > 0: \mathbb{E}[\exp(|X|/t)] \leq 2\},
    \end{equation}
    cf. Definition 2.7.5 of \cite{vershynin_high-dimensional_2018}. We call a random variable for which $\| \cdot \|_{\psi_1}$ is finite sub-exponential, following, e.g., \cite{vershynin_high-dimensional_2018}.

    Define now $Z_i:= X_i Y_i - (\muu_x)_i (\muu_y)_i$. We observe that
    \[
        Z_i = X_i Y_i - (\muu_x)_i (\muu_y)_i = X_i (Y_i - (\muu_y)_i) + (X_i - (\muu_x)_i) (\muu_y)_i = Z_i^{(1)} + Z_i^{(2)}
    \]
    with $Z_i^{(1)}:= X_i (Y_i - (\muu_y)_i)$ and $Z_i^{(2)}:= (X_i - (\muu_x)_i) (\muu_y)_i$. To bound the $\psi_1$-norm of $Z_i$, we bound this norm for $Z_i^{(1)}$ and $Z_i^{(2)}$ separately.

    Indeed, due to Lemma 2.7.7 of \cite{vershynin_high-dimensional_2018}, it holds that
    \[
        \|Z_i^{(1)}\|_{\psi_1} \leq \| X_i\|_{\psi_2} \|Y_i - (\muu_y)_i\|_{\psi_2},
    \]
    where
    \begin{equation} \label{eq:def:psi2}
        \| X\|_{\psi_2} := \inf \{t > 0: \mathbb{E}[\exp(X^2/t^2)] \leq 2\},
    \end{equation}
    is the $\psi_2$-Orlicz norm \cite{vershynin_high-dimensional_2018} characterizing sub-Gaussian random variables $X$. From Lemma \ref{lemma:psi2:Gaussian}, it follows therefore that
    \[
        \|Z_i^{(1)}\|_{\psi_1} \leq \max\left( 2 (\sig_x)_i, \sqrt{\frac{(\muu_x)_i^2}{\log 2}+ (\sig_x)_i^2 }\right) \sqrt{\frac{8}{3}} (\sig_y)_i.
    \]
    For the second part, since $\| \cdot \|_{\psi_2}$ is a norm, we estimate that
    \[
        \|Z_i^{(2)}\|_{\psi_1} \leq \|X_i - (\muu_x)_i \|_{\psi_2} \|(\muu_y)_i\|_{\psi_2} \leq \sqrt{\frac{8}{3}} (\sig_x)_i \|(\muu_y)_i\|_{\psi_2} \leq  \sqrt{\frac{8}{3}} (\sig_x)_i \frac{|(\muu_y)_i|}{\sqrt{\log 2}} ,
    \]
    where we used again Lemma 2.7.7 and (2.17) of \cite{vershynin_high-dimensional_2018} in the first and last inequality, respectively, and Lemma \ref{lemma:psi2:Gaussian} in the second inequality.

    From this, it follows that
    \begin{equation} \label{eq:bound:Zi:psi1}
        \begin{split}
            \|Z_i\|_{\psi_1} \leq \|Z_i^{(1)} \|_{\psi_1} + \|Z_i^{(2)} \|_{\psi_1}
             & \leq  \max\left( 2 (\sig_x)_i, \sqrt{\frac{(\muu_x)_i^2}{\log 2}+ (\sig_x)_i^2 }\right) \sqrt{\frac{8}{3}} (\sig_y)_i + \sqrt{\frac{8}{3}} (\sig_x)_i \frac{|(\muu_y)_i|}{\sqrt{\log 2}} \\
             & \leq \left(2 (\sig_x)_i + \sqrt{\frac{(\muu_x)_i^2}{\log 2}+ (\sig_x)_i^2 }\right) \sqrt{\frac{8}{3}} (\sig_y)_i + \sqrt{\frac{8}{3}} (\sig_x)_i \frac{|(\muu_y)_i|}{\sqrt{\log 2}}      \\
             & \leq 2 \sqrt{6} (\sig_x)_i (\sig_y)_i + \sqrt{\frac{8}{3}} (\sig_y)_i \frac{|(\muu_x)_i|}{\sqrt{\log 2}}  + \sqrt{\frac{8}{3}} (\sig_x)_i \frac{|(\muu_y)_i|}{\sqrt{\log 2}},
        \end{split}
    \end{equation}
    using that $\sqrt{a^2 + b^2} \leq a + b$ for any non-negative $a,b\geq 0$ in the last inequality. We next establish a lower bound on the probability of a sign flip through normalization, i.e., for
    $
        \Pr(\sign\left(\mathbf{x}^\top\mathbf{y}\right)\ne\sign\left(\tilde{\mathbf{x}}^\top\tilde{\mathbf{y}}\right))$. Assuming without loss of generality that $|\muu_x^{\top} \muu_y| = \muu_x^{\top} \muu_y$, we observe that
    \[
        \begin{split}
            \Pr\left(\sign\left(\mathbf{x}^\top\mathbf{y}\right)\ne\sign\left(\tilde{\mathbf{x}}^\top\tilde{\mathbf{y}}\right)\right) & = \Pr\left( (\mathbf{x}^\top\mathbf{y}  > 0) \wedge (\tilde{\mathbf{x}}^\top\tilde{\mathbf{y}} < 0) \right) + \Pr\left( (\mathbf{x}^\top\mathbf{y}  < 0) \wedge (\tilde{\mathbf{x}}^\top\tilde{\mathbf{y}}  > 0) \right) \\
                                                                                                                                      & \geq \Pr\left( (\mathbf{x}^\top\mathbf{y}  > 0) \wedge (\tilde{\mathbf{x}}^\top\tilde{\mathbf{y}} < 0) \right).
        \end{split}
    \]
    Furthermore, since the distribution of the normalized vectors $\tilde{\mathbf{x}}$ and $\tilde{\mathbf{y}}$ is symmetric, the same holds true for the dot product $\tilde{\mathbf{x}}^\top\tilde{\mathbf{y}}$, which implies that
    \[
        \begin{split}
            \Pr\left( (\mathbf{x}^\top\mathbf{y}  > 0) \wedge (\tilde{\mathbf{x}}^\top\tilde{\mathbf{y}} < 0) \right) & = 1 - \Pr\left( (\mathbf{x}^\top\mathbf{y}  \leq 0) \vee  (\tilde{\mathbf{x}}^\top\tilde{\mathbf{y}} \geq 0) \right) \geq 1 -  \Pr\left( \mathbf{x}^\top\mathbf{y}  \leq 0 \right) - \Pr\left(\tilde{\mathbf{x}}^\top\tilde{\mathbf{y}} \geq 0 \right) \\
                                                                                                                      & \geq 1 - 0.5 - \Pr\left( \mathbf{x}^\top\mathbf{y}  \leq 0 \right) = 0.5 - \Pr\left( \mathbf{x}^\top\mathbf{y}  \leq 0 \right).
        \end{split}
    \]
    It remains to show that
    \begin{equation} \label{eq:prob:bound}
        \Pr\left( \mathbf{x}^\top\mathbf{y}  \leq 0 \right) \leq 0.1.
    \end{equation}
    To establish this, we see that
    \[
        \begin{split}
            \Pr\left( \mathbf{x}^\top\mathbf{y}  \leq 0 \right) & = \Pr\left(\mathbf{x}^\top\mathbf{y} - \mathbb{E}\left[\mathbf{x}^\top\mathbf{y}\right] \leq -\mathbb{E}\left[\mathbf{x}^\top\mathbf{y}\right]\right) = \Pr\left(\mathbf{x}^\top\mathbf{y} - \boldsymbol{\mu}_x^\top\boldsymbol{\mu}_y  \leq -\boldsymbol{\mu}_x^\top\boldsymbol{\mu}_y \right) \\
                                                                & = \Pr\left( \sum_{i=1}^D Z_i \leq -\boldsymbol{\mu}_x^\top\boldsymbol{\mu}_y \right)
        \end{split}
    \]
    with the random variables $Z_i$ defined above. Using the triangle inequality of the $\ell_2$-norm, it follows from \eqref{eq:bound:Zi:psi1} that
    \[
        \sqrt{\sum_{i=1}^D \|Z_i\|_{\psi_1}^2} \leq 2 \sqrt{6} \sqrt{ (\sig_x^2)^{\top} \sig_y^2} + \sqrt{\frac{8}{3 \log 2}}\left( \sqrt{(\sig_y^2)^{\top} \muu_x^2} + \sqrt{(\sig_x^2)^{\top} \muu_y^2} \right)
    \]
    and that
    \[
        \max_{i=1}^{D} \|Z_i\|_{\psi_1} \leq  2 \sqrt{6} \| \sig_x \circ \sig_y \|_{\infty} + \sqrt{\frac{8}{3 \log 2}} \left( \| \sig_y \circ |(\muu_x)| \|_{\infty} + \| \sig_x \circ  |(\muu_y)| \|_{\infty} \right),
    \]
    which implies that
    \[
        \begin{split}
             & \sqrt{2 \sum_{i=1}^D \|Z_i\|_{\psi_1}^2} \sqrt{\log(10)} + \max_{i=1}^{D} \|Z_i\|_{\psi_1} \log(10)                                                                                                                                                                         \\
             & \leq 4 \sqrt{3 \log(10)} \sqrt{ (\sig_x^2)^{\top} \sig_y^2} + \frac{4 \sqrt{\log(10)}}{\sqrt{3 \log(2)}} \left( \sqrt{(\sig_y^2)^{\top} \muu_x^2} + \sqrt{(\sig_x^2)^{\top} \muu_y^2} \right)                                                                               \\
             & + 2 \sqrt{6}\log(10) \| \sig_x \circ \sig_y \|_{\infty} + \sqrt{\frac{8}{3 \log 2}} \log(10) \left( \| \sig_y \circ |(\muu_x)| \|_{\infty} + \| \sig_x \circ  |(\muu_y)| \|_{\infty} \right)                                                                                \\
             & \leq 12 \left(\sqrt{ (\sig_x^2)^{\top} \sig_y^2} +  \| \sig_x \circ \sig_y \|_{\infty} \right)+ 5  \left( \sqrt{(\sig_y^2)^{\top} \muu_x^2} + \sqrt{(\sig_x^2)^{\top} \muu_y^2}+ \| \sig_y \circ |(\muu_x)| \|_{\infty} + \| \sig_x \circ  |(\muu_y)| \|_{\infty}   \right) \\
             & \leq  |\muu_{x}^{\top} \muu_{y}|,
        \end{split}
    \]
    using the assumption \eqref{eq:signflip:theorem:assumption} in the last inequality. With this inequality, we can use the fact that the $Z_i$ are independent mean-zero sub-exponential random variables and Bernstein's inequality as stated in Lemma \ref{lemma:Bernstein} to conclude that
    \[
        \begin{split}
            \Pr\left( \sum_{i=1}^D Z_i \leq -\boldsymbol{\mu}_x^\top\boldsymbol{\mu}_y \right) & \leq \Pr\left( \sum_{i=1}^D Z_i \leq -\left(\sqrt{2 \sum_{i=1}^D \|Z_i\|_{\psi_1}^2} \sqrt{\log(10)} + \max_{i=1}^{D} \|Z_i\|_{\psi_1} \log(10) \right) \right)
            \\
                                                                                               & \leq \exp(-\log(10)) = 0.1.
        \end{split}
    \]
    This establishes \eqref{eq:prob:bound}, which concludes the proof. \end{proof}

\begin{lemma}[{Bernstein's Inequality, cf. Lemma 5.1 of \cite{dirksen_tail_2015}}]
    \label{lemma:Bernstein}
    Let $Z_1,\ldots Z_D$ be independent mean-zero sub-exponential random variables. Then for every $t \geq 0$,
    \[
        \Pr \left(  \sum_{i=1}^D Z_i   \leq -\left( \sqrt{2 \sum_{i=1}^D \|Z_i\|_{\psi_1}^2} \sqrt{t} + \max_{i=1}^{D} \|Z_i\|_{\psi_1} t \right) \right) \leq \exp(-t).
    \]
\end{lemma}

\begin{lemma}[{Bounds on $\psi_2$-norm of Gaussians \cite{vershynin_high-dimensional_2018}}] \label{lemma:psi2:Gaussian}
    \begin{enumerate}
        \item If $X \sim \mathcal{N}(0,\sigma^2)$ is a centered Gaussian random variable with variance $\sigma^2$, then its $\psi_2$-norm \eqref{eq:def:psi2} satisfies
              \[
                  \|X\|_{\psi_2} \leq \sqrt{\frac{8}{3}} \sigma.
              \]
        \item If $X \sim \mathcal{N}(\mu,\sigma^2)$ is a Gaussian random variable with mean $\mu$ and variance $\sigma^2$, then its $\psi_2$-norm \eqref{eq:def:psi2} satisfies
              \[
                  \|X\|_{\psi_2} \leq \max\Big( 2 \sigma, \sqrt{\frac{\mu^2}{\log 2}+ \sigma^2 }\Big).
              \]
    \end{enumerate}
\end{lemma}

\subsection{Proof of \cref{thm:unit-norm-gradient}}

\begin{proof}[Proof of \cref{thm:unit-norm-gradient}]
    Given the output of an affine transformation $\mathbf{x} = \mathbf{W}\mathbf{v} + \mathbf{b}$, where $\mathbf{W}$ and $\mathbf{b}$ are learnable parameters. If $\mathbf{x}' = (\alpha\mathbf{W})\mathbf{v} + (\alpha\mathbf{b})$, then the output of UnitNorm is unchanged, \ie, $\tilde{\mathbf{x}}' = \tilde{\mathbf{x}}$, while the gradients to loss $\mathcal{L}$ are given as follows:
    \begin{equation}
        \begin{aligned}
            \frac{\partial \mathcal{L}}{\partial \tilde{\mathbf{x}}'} \cdot \frac{\partial \tilde{\mathbf{x}}'}{\partial (\alpha\mathbf{W})}
             & = \frac{1}{\alpha} \cdot \frac{\partial \mathcal{L}}{\partial \tilde{\mathbf{x}}} \cdot \frac{\partial \tilde{\mathbf{x}}}{\partial \mathbf{W}}
             &                                                                                                                                                 & = \frac{1}{\alpha} \cdot \frac{\partial \mathcal{L}}{\partial \tilde{\mathbf{x}}} \cdot \mathbf{J} \mathbf{v}^\top \\
            \frac{\partial \mathcal{L}}{\partial \tilde{\mathbf{x}}'} \cdot \frac{\partial \tilde{\mathbf{x}}'}{\partial (\alpha\mathbf{b})}
             & = \frac{1}{\alpha} \cdot \frac{\partial \mathcal{L}}{\partial \tilde{\mathbf{x}}} \cdot \frac{\partial \tilde{\mathbf{x}}}{\partial \mathbf{b}}
             &                                                                                                                                                 & = \frac{1}{\alpha} \cdot \frac{\partial \mathcal{L}}{\partial \tilde{\mathbf{x}}} \cdot \mathbf{J}                 \\
            \frac{\partial \mathcal{L}}{\partial \tilde{\mathbf{x}}'} \cdot \frac{\partial \tilde{\mathbf{x}}'}{\partial \mathbf{v}}
             & = \frac{\partial \mathcal{L}}{\partial \tilde{\mathbf{x}}} \cdot \frac{\partial \tilde{\mathbf{x}}}{\partial \mathbf{v}}
             &                                                                                                                                                 & = \frac{\partial \mathcal{L}}{\partial \tilde{\mathbf{x}}} \cdot \mathbf{J} \mathbf{W}^\top
        \end{aligned}
    \end{equation}
    Proof: First we will show $\tilde{\mathbf{x}}' = \tilde{\mathbf{x}}$, for which we have:
    \begin{equation}
        \begin{aligned}
            \tilde{\mathbf{x}}' = & D^{\frac{k}{2}}\frac{\mathbf{x}'}{\left\|\mathbf{x}'\right\|}           \\
            =                     & D^{\frac{k}{2}}\frac{\alpha\mathbf{x}}{\alpha\left\|\mathbf{x}\right\|} \\
            =                     & D^{\frac{k}{2}}\frac{\mathbf{x}}{\left\|\mathbf{x}\right\|}             \\
            =                     & \tilde{\mathbf{x}}
        \end{aligned}
    \end{equation}
    And thus for the gradients to loss $\mathcal{L}$, we have $\frac{\partial \mathcal{L}}{\partial \tilde{\mathbf{x}}'} = \frac{\partial \mathcal{L}}{\partial \tilde{\mathbf{x}}}$. Also, for the Jacobian matrix $\mathbf{J}$ of $\tilde{\mathbf{x}}$ \wrt $\mathbf{x}$, we have
    \begin{equation}
        \begin{aligned}
            \mathbf{J} = & \frac{\partial D^{\frac{k}{2}}\frac{\mathbf{x}}{\left\|\mathbf{x}\right\|}}{\partial \mathbf{x}}                                         \\
            =            & D^{\frac{k}{2}}\left(\frac{\mathbf{I}}{\left\|\mathbf{x}\right\|} - \frac{\mathbf{x}\mathbf{x}^\top}{\left\|\mathbf{x}\right\|^3}\right)
        \end{aligned}
    \end{equation}
    And the Jacobian matrix of $\tilde{\mathbf{x}}'$ \wrt $\mathbf{x}'$ is given as:
    \begin{equation}
        \begin{aligned}
            \frac{\partial D^{\frac{k}{2}}\frac{\mathbf{x}'}{\left\|\mathbf{x}'\right\|}}{\partial \mathbf{x}'} = & D^{\frac{k}{2}}\left(\frac{\mathbf{I}}{\left\|\mathbf{x}'\right\|} - \frac{\mathbf{x}'\mathbf{x}'^\top}{\left\|\mathbf{x}'\right\|^3}\right)                   \\
            =                                                                                                     & D^{\frac{k}{2}}\left(\frac{\mathbf{I}}{\alpha\left\|\mathbf{x}\right\|} - \frac{\alpha^2\mathbf{x}\mathbf{x}^\top}{\alpha^3\left\|\mathbf{x}\right\|^3}\right) \\
            =                                                                                                     & \frac{1}{\alpha} D^{\frac{k}{2}}\left(\frac{\mathbf{I}}{\left\|\mathbf{x}\right\|} - \frac{\mathbf{x}\mathbf{x}^\top}{\left\|\mathbf{x}\right\|^3}\right)      \\
            =                                                                                                     & \frac{1}{\alpha}\mathbf{J}
        \end{aligned}
    \end{equation}
    Then we have the gradient of loss \wrt $\mathbf{W}$ and $\alpha\mathbf{W}$:
    \begin{equation}
        \begin{aligned}
            \frac{\partial \mathcal{L}}{\partial \tilde{\mathbf{x}}} \cdot \frac{\partial \tilde{\mathbf{x}}}{\partial \mathbf{W}} =                       & \frac{\partial \mathcal{L}}{\partial \tilde{\mathbf{x}}} \cdot \frac{\partial \tilde{\mathbf{x}}}{\partial \mathbf{x}} \cdot \frac{\partial \mathbf{x}}{\partial \mathbf{W}}                                                                                    \\
            =                                                                                                                                              & \frac{\partial \mathcal{L}}{\partial \tilde{\mathbf{x}}} \cdot \mathbf{J} \mathbf{v}^\top                                                                                                                                                                       \\
            \frac{\partial \mathcal{L}}{\partial \tilde{\mathbf{x}}'} \cdot \frac{\partial \tilde{\mathbf{x}}'}{\partial (\alpha\mathbf{W})} =             & \frac{\partial \mathcal{L}}{\partial \tilde{\mathbf{x}}'} \cdot \frac{\partial \tilde{\mathbf{x}}'}{\partial \mathbf{x}'} \cdot \frac{\partial \mathbf{x}'}{\partial (\alpha\mathbf{W})}                                                                        \\
            =                                                                                                                                              & \frac{\partial \mathcal{L}}{\partial \tilde{\mathbf{x}}} \cdot \frac{1}{\alpha} \mathbf{J} \mathbf{v}^\top                                                                                                                                                      \\
            \Rightarrow \frac{\partial \mathcal{L}}{\partial \tilde{\mathbf{x}}'} \cdot \frac{\partial \tilde{\mathbf{x}}'}{\partial (\alpha\mathbf{W})} = & \frac{1}{\alpha} \cdot \frac{\partial \mathcal{L}}{\partial \tilde{\mathbf{x}}} \cdot \frac{\partial \tilde{\mathbf{x}}}{\partial \mathbf{W}} = \frac{1}{\alpha} \cdot \frac{\partial \mathcal{L}}{\partial \tilde{\mathbf{x}}} \cdot \mathbf{J}\mathbf{v}^\top
        \end{aligned}
    \end{equation}
    Similarly, for $\mathbf{b}$ and $\alpha\mathbf{b}$ we have:
    \begin{equation}
        \begin{aligned}
            \frac{\partial \mathcal{L}}{\partial \tilde{\mathbf{x}}} \cdot \frac{\partial \tilde{\mathbf{x}}}{ \partial \mathbf{b}} =                       & \frac{\partial \mathcal{L}}{\partial \tilde{\mathbf{x}}} \cdot \frac{\partial \tilde{\mathbf{x}}}{\partial \mathbf{x}} \cdot \frac{\partial \mathbf{x}}{\partial \mathbf{b}}                                                                     \\
            =                                                                                                                                               & \frac{\partial \mathcal{L}}{\partial \tilde{\mathbf{x}}} \cdot \mathbf{J}                                                                                                                                                                        \\
            \frac{\partial \mathcal{L}}{\partial \tilde{\mathbf{x}}'} \cdot \frac{\partial \tilde{\mathbf{x}}'}{ \partial (\alpha\mathbf{b})} =             & \frac{\partial \mathcal{L}}{\partial \tilde{\mathbf{x}}'} \cdot \frac{\partial \tilde{\mathbf{x}}'}{\partial \mathbf{x}'} \cdot \frac{\partial \mathbf{x}'}{\partial (\alpha\mathbf{b})}                                                         \\
            =                                                                                                                                               & \frac{\partial \mathcal{L}}{\partial \tilde{\mathbf{x}}} \cdot \frac{1}{\alpha}\mathbf{J}                                                                                                                                                        \\
            \Rightarrow \frac{\partial \mathcal{L}}{\partial \tilde{\mathbf{x}}'} \cdot \frac{\partial \tilde{\mathbf{x}}'}{ \partial (\alpha\mathbf{b})} = & \frac{1}{\alpha} \cdot \frac{\partial \mathcal{L}}{\partial \tilde{\mathbf{x}}} \cdot \frac{\partial \tilde{\mathbf{x}}}{\partial \mathbf{b}} = \frac{1}{\alpha} \cdot \frac{\partial \mathcal{L}}{\partial \tilde{\mathbf{x}}} \cdot \mathbf{J}
        \end{aligned}
    \end{equation}
    And for $\mathbf{v}$, we have:
    \begin{equation}
        \begin{aligned}
            \frac{\partial \mathcal{L}}{\partial \tilde{\mathbf{x}}} \cdot \frac{\partial \tilde{\mathbf{x}}}{ \partial \mathbf{v}} =               & \frac{\partial \mathcal{L}}{\partial \tilde{\mathbf{x}}} \cdot \frac{\partial \tilde{\mathbf{x}}}{\partial \mathbf{x}} \cdot \frac{\partial \mathbf{x}}{\partial \mathbf{v}}                                       \\
            =                                                                                                                                       & \frac{\partial \mathcal{L}}{\partial \tilde{\mathbf{x}}} \cdot \mathbf{J} \mathbf{W}^\top                                                                                                                          \\
            \frac{\partial \mathcal{L}}{\partial \tilde{\mathbf{x}}'} \cdot \frac{\partial \tilde{\mathbf{x}}'}{ \partial \mathbf{v}} =             & \frac{\partial \mathcal{L}}{\partial \tilde{\mathbf{x}}'} \cdot \frac{\partial \tilde{\mathbf{x}}'}{\partial \mathbf{x}'} \cdot \frac{\partial \mathbf{x}'}{\partial \mathbf{v}}                                   \\
            =                                                                                                                                       & \frac{\partial \mathcal{L}}{\partial \tilde{\mathbf{x}}} \cdot \frac{1}{\alpha}\mathbf{J} (\alpha\mathbf{W})^\top                                                                                                  \\
            \Rightarrow \frac{\partial \mathcal{L}}{\partial \tilde{\mathbf{x}}'} \cdot \frac{\partial \tilde{\mathbf{x}}'}{ \partial \mathbf{v}} = & \frac{\partial \mathcal{L}}{\partial \tilde{\mathbf{x}}} \cdot \frac{\partial \tilde{\mathbf{x}}}{\partial \mathbf{v}} = \frac{\partial \mathcal{L}}{\partial \tilde{\mathbf{x}}} \cdot \mathbf{J} \mathbf{W}^\top
        \end{aligned}
    \end{equation}
\end{proof}

\subsection{Proofs of \cref{thm:unit-norm-entropy} and \cref{lem:unit-norm-entropy-limit}}
\begin{proof}[Proof of \cref{thm:unit-norm-entropy}]
    Let $\mathbf{X} \in \mathbb{R}^{L \times D}$ be a single sequence of token vectors, and let $\tilde{\mathbf{X}}$ be the unit normalized output with modulus $k$, the entropy lower bound (ELB) of the attention scores is given by the following expression:
    \begin{equation}
        \begin{aligned}
            \ELB(k;L,D) = & \min_{i=1}^L H(\mathbf{A}_i)                                                                                                                                     \\
            =             & \min_{i=1}^L \left(-\sum_{j=1}^{L}\mathbf{A}_{i,j}\log\mathbf{A}_{i,j}\right)                                                                                    \\
            =             & \log\left(L-1+\exp\left(2D^{k-\frac{1}{2}}\right)\right) - \frac{2D^{k-\frac{1}{2}}\exp\left(2D^{k-\frac{1}{2}}\right)}{L-1+\exp\left(2D^{k-\frac{1}{2}}\right)}
        \end{aligned}
    \end{equation}
    Proof: Let $\tilde{\mathbf{X}} = D^{\frac{k}{2}}\mathbf{e}$ where $\mathbf{e}$ are the vectors of unit norm. Without loss of generality, we can assume the ELB is achieved at anchor index $i$, where we can compute the attention scores as follows:
    \begin{equation}
        \begin{aligned}
            \mathbf{A}_i = & \softmax\left(\frac{\tilde{\mathbf{X}}_i\tilde{\mathbf{X}}^\top}{\sqrt{D}}\right) \\
            =              & \softmax\left(\frac{D^k \mathbf{e}_i\mathbf{e}^\top}{\sqrt{D}}\right)             \\
            =              & \softmax\left(D^{k-\frac{1}{2}}\mathbf{e}_i\mathbf{e}^\top\right)
        \end{aligned}
    \end{equation}
    Since $\mathbf{e}_i\mathbf{e}_j^\top\in\left(-1,1\right),\forall i,j = 1,2,\cdots,L$, the entropy of the attentions scores is lower bounded by the following expression when it satisfies that $\mathbf{e}_i\mathbf{e}_j = \begin{cases}
            1,       & j=i     \\
            \RB{-1}, & j\neq i
        \end{cases}$:
    \begin{equation}
        \begin{aligned}
              & H(\mathbf{A}_i)                                                                                                                                                                                                                                                 \\
            = & -\sum_{j=1}^{L}\mathbf{A}_{i,j}\log\mathbf{A}_{i,j}                                                                                                                                                                                                             \\
            = & -(L-1) \cdot \frac{\exp\left(-D^{k-\frac{1}{2}}\right)}{(L-1)\exp\left(-D^{k-\frac{1}{2}}\right)+\exp\left(D^{k-\frac{1}{2}}\right)}\log\frac{\exp\left(-D^{k-\frac{1}{2}}\right)}{(L-1)\exp\left(-D^{k-\frac{1}{2}}\right)+\exp\left(D^{k-\frac{1}{2}}\right)} \\
              & - \frac{\exp\left(D^{k-\frac{1}{2}}\right)}{(L-1)\exp\left(-D^{k-\frac{1}{2}}\right)+\exp\left(D^{k-\frac{1}{2}}\right)}\log \frac{\exp\left(D^{k-\frac{1}{2}}\right)}{(L-1)\exp\left(-D^{k-\frac{1}{2}}\right)+\exp\left(D^{k-\frac{1}{2}}\right)}             \\
            = & \frac{L-1}{L-1+\exp\left(2D^{k-\frac{1}{2}}\right)}\log\left(L-1+\exp\left(2D^{k-\frac{1}{2}}\right)\right)                                                                                                                                                     \\
              & + \frac{\exp\left(2D^{k-\frac{1}{2}}\right)}{L-1+\exp\left(2D^{k-\frac{1}{2}}\right)}\log\frac{L-1+\exp\left(2D^{k-\frac{1}{2}}\right)}{\exp\left(2D^{k-\frac{1}{2}}\right)}                                                                                    \\
            = & \log\left(L-1+\exp\left(2D^{k-\frac{1}{2}}\right)\right) - \frac{2D^{k-\frac{1}{2}}\exp\left(2D^{k-\frac{1}{2}}\right)}{L-1+\exp\left(2D^{k-\frac{1}{2}}\right)}
        \end{aligned}
    \end{equation}
    Therefore, the entropy lower bound (ELB) for any $L,D$ and $k$ is:
    \begin{equation}
        \ELB(k;L,D) = \log\left(L-1+\exp\left(2D^{k-\frac{1}{2}}\right)\right) - \frac{2D^{k-\frac{1}{2}}\exp\left(2D^{k-\frac{1}{2}}\right)}{L-1+\exp\left(2D^{k-\frac{1}{2}}\right)}
    \end{equation}
\end{proof}

\begin{proof}[Proof of \cref{lem:unit-norm-entropy-limit}]
    The ELB is a monotonically decreasing function of $k$ bounded between $0$ and $\log{L}$.

    Proof: Let $d = 2D^{k-\frac{1}{2}}$, then it is obvious that $d$ is monotonically increasing with $k$, therefore we only need to prove that $\ELB(k;L,D)$ is monotonically decreasing with $d$. The derivative of $\ELB(k;L,D)$ with respect to $d$ is given as follows:
    \begin{equation}
        \begin{aligned}
            \frac{\partial \ELB(k;L,D)}{\partial d} = & \frac{e^d}{L-1+e^d} -  \frac{\left(L-1+e^d\right)(d+1)e^d - (de^d)e^d}{\left(L-1+e^d\right)^2}         \\
            =                                         & \frac{e^d}{\left(L-1+e^d\right)^2}\left(\left(L-1+e^d\right) - \left(L-1+e^d\right)(d+1) + de^d\right) \\
            =                                         & \frac{de^d}{\left(L-1+e^d\right)^2}(1-L)                                                               \\
            (\forall L>1) <                           & 0                                                                                                      \\
        \end{aligned}
    \end{equation}
    Therefore, $\ELB(k;L,D)$ is monotonically decreasing with $d$ and with $k$. If the limits of $\ELB(k;L,D)$ as $k\to-\infty$ and $k\to+\infty$ exist, then $\ELB(k;L,D)$ is bounded between these two limits. The limits are given as follows:
    \begin{equation}
        \begin{aligned}
            \lim_{k\to-\infty} \ELB(k;L,D) = & \lim_{d\to 0^+} \left(\log\left(L-1+e^d\right) - \frac{de^d}{L-1+e^d}\right) \\
            =                                & \log\left(L-1+1\right) - \frac{0}{L-1+1}                                     \\
            =                                & \log{L}                                                                      \\
        \end{aligned}
    \end{equation}
    \begin{equation}
        \begin{aligned}
            \lim_{k\to+\infty} \ELB(k;L,D) = & \lim_{d\to+\infty} \left(\log\left(L-1+e^d\right) - \frac{de^d}{L-1+e^d}\right) \\
            =                                & \lim_{d\to+\infty} \log e^d - \lim_{d\to+\infty} \frac{d}{(L-1)e^{-d}+1}        \\
            =                                & d - d                                                                           \\
            =                                & 0                                                                               \\
        \end{aligned}
    \end{equation}
    Therefore, $\ELB(k;L,D)$ is bounded between $0$ and $\log{L}$.
\end{proof}

%% file: 930.appendix.discussion.tex
\section{Discussion}

\subsection{Difference between the proposed normalization and the other normalization} \label{sec:norm-difference}

BatchNorm and LayerNorm are all normalization methods that are widely used in deep learning. They share the same center-and-scale normalization paradigm by first subtracting the mean and then divide by standard deviation. The only difference between them in terms of computation is the dimensions of data used to compute these statistics, as shown in \cref{tab:norm-difference}.

In terms of application, BatchNorm is often used in fully connected layers and convolution layers, while LayerNorm is often used in recurrent neural networks and Transformers. The subtle difference between LayerNorm (theory) and LayerNorm (practice) might be attributed to the fact that the sequence length $L$ is often variable in Transformers, thus normalization within each token might be more stable. But this will require further investigation to come to a conclusion.

The proposed UnitNorm is a normalization method that is used to normalize the input data to have unit norm, which takes the same dimension for computation as LayerNorm, yet it distinguishes itself from LayerNorm by the fact that it does not subtract the mean and divide by standard deviation. Also, UnitNorm discard the center operation on the normalized output, as it will also cause the problem of token shift (\cref{sec:token-shift}).

\begin{table}[!htb]
    \caption{Computation of the statistics for different normalization methods. Input data $\mathbf{X} \in \mathbb{R}^{N \times L \times D}$, where $N$ is the batch size, $L$ is the sequence length and $D$ is the feature dimension. $\mathbf{X}_{n,l,d}$ denotes the $d$-th feature of the $l$-th token in the $n$-th sequence. Normalization is broadcasted over the same dimension as the statistics and mathematical operations are done element-wise. For BatchNorm, LayerNorm (theory) and LayerNorm (practice), $\boldsymbol{\gamma}$ and $\boldsymbol{\beta}$ are optional learnable parameters that will re-scale and re-center the normalized output element-wise, which is enabled by default in the PyTorch's implementation.}
    \label{tab:norm-difference}
    \vskip 0.15in
    \begin{center}
        \begin{small}
            \begin{tabular}{c|l|l}
                \toprule
                Method               & Statistics               & Normalization \\
                \midrule
                BatchNorm            & $
                    \begin{aligned}
                        \mu_d =                 & \frac{1}{NL}\sum_{n=1}^{N}\sum_{l=1}^{L}\mathbf{X}_{n,l,d}           \\
                        \sigma_d^2 =            & \frac{1}{NL}\sum_{n=1}^{N}\sum_{l=1}^{L}(\mathbf{X}_{n,l,d}-\mu_d)^2 \\
                        \boldsymbol{\mu} =      & \begin{bmatrix}
                                                      \mu_1 & \mu_2 & \cdots & \mu_D
                                                  \end{bmatrix}^\top \in \mathbb{R}^{1 \times 1 \times D}              \\
                        \boldsymbol{\sigma}^2 = & \begin{bmatrix}
                                                      \sigma_1^2 & \sigma_2^2 & \cdots & \sigma_D^2
                                                  \end{bmatrix}^\top \in \mathbb{R}^{1 \times 1 \times D}
                    \end{aligned}
                $                    & \multirow{3}{*}[-32ex]{$
                        \begin{aligned}
                            \tilde{\mathbf{X}} = & \frac{\mathbf{X} - \boldsymbol{\mu}}{\sqrt{\boldsymbol{\sigma}^2 + \varepsilon}} \\
                            \mathbf{Y} =         & \tilde{\mathbf{X}} \odot \boldsymbol{\gamma} + \boldsymbol{\beta}
                        \end{aligned}
                $}                                                              \\
                \cmidrule{1-2}
                LayerNorm (theory)   & $
                    \begin{aligned}
                        \mu_n =                 & \frac{1}{LD}\sum_{l=1}^{L}\sum_{d=1}^{D}\mathbf{X}_{n,l,d}           \\
                        \sigma_n^2 =            & \frac{1}{LD}\sum_{l=1}^{L}\sum_{d=1}^{D}(\mathbf{X}_{n,l,d}-\mu_n)^2 \\
                        \boldsymbol{\mu} =      & \begin{bmatrix}
                                                      \mu_1 & \mu_2 & \cdots & \mu_N
                                                  \end{bmatrix}^\top \in \mathbb{R}^{N \times 1 \times 1}              \\
                        \boldsymbol{\sigma}^2 = & \begin{bmatrix}
                                                      \sigma_1^2 & \sigma_2^2 & \cdots & \sigma_N^2
                                                  \end{bmatrix}^\top \in \mathbb{R}^{N \times 1 \times 1}
                    \end{aligned}
                $                    &                                          \\
                \cmidrule{1-2}
                LayerNorm (practice) & $
                    \begin{aligned}
                        \mu_{n,l} =             & \frac{1}{D}\sum_{d=1}^{D}\mathbf{X}_{n,l,d}               \\
                        \sigma_{n,l}^2 =        & \frac{1}{D}\sum_{d=1}^{D}(\mathbf{X}_{n,l,d}-\mu_n)^2     \\
                        \boldsymbol{\mu} =      & \begin{bmatrix}
                                                      \mu_{1,1} & \mu_{1,2} & \cdots & \mu_{1,L} \\
                                                      \mu_{2,1} & \mu_{2,2} & \cdots & \mu_{2,L} \\
                                                      \vdots    & \vdots    & \ddots & \vdots    \\
                                                      \mu_{N,1} & \mu_{N,2} & \cdots & \mu_{N,L}
                                                  \end{bmatrix}^\top \in \mathbb{R}^{N \times L \times 1}   \\
                        \boldsymbol{\sigma}^2 = & \begin{bmatrix}
                                                      \sigma_{1,1}^2 & \sigma_{1,2}^2 & \cdots & \sigma_{1,L}^2 \\
                                                      \sigma_{2,1}^2 & \sigma_{2,2}^2 & \cdots & \sigma_{2,L}^2 \\
                                                      \vdots         & \vdots         & \ddots & \vdots         \\
                                                      \sigma_{N,1}^2 & \sigma_{N,2}^2 & \cdots & \sigma_{N,L}^2
                                                  \end{bmatrix}^\top \in \mathbb{R}^{N \times L \times 1}
                    \end{aligned}
                $                    &                                          \\
                \midrule
                \makecell{\vspace{3ex}                                          \\ RMSNorm \\\vspace{3ex}} & \multirow{2}{*}[4ex]{$
                        \begin{aligned}
                            \left\|\mathbf{X}\right\|_{n,l} = & \sqrt{\sum_{d=1}^{D}\mathbf{X}_{n,l,d}^2}                                                                    \\
                            \left\|\mathbf{X}\right\| =       & \begin{bmatrix}
                                                                    \left\|\mathbf{X}\right\|_{1,1} & \left\|\mathbf{X}\right\|_{1,2} & \cdots & \left\|\mathbf{X}\right\|_{1,L} \\
                                                                    \left\|\mathbf{X}\right\|_{2,1} & \left\|\mathbf{X}\right\|_{2,2} & \cdots & \left\|\mathbf{X}\right\|_{2,L} \\
                                                                    \vdots                          & \vdots                          & \ddots & \vdots                          \\
                                                                    \left\|\mathbf{X}\right\|_{N,1} & \left\|\mathbf{X}\right\|_{N,2} & \cdots & \left\|\mathbf{X}\right\|_{N,L}
                                                                \end{bmatrix}^\top \\
                            \in                               & \mathbb{R}^{N \times L \times 1}
                        \end{aligned}
                $}                   & $
                    \begin{aligned}
                        \tilde{\mathbf{X}} = & \sqrt{D} \frac{\mathbf{X}}{\left\|\mathbf{X}\right\|}             \\
                        \mathbf{Y} =         & \tilde{\mathbf{X}} \odot \boldsymbol{\gamma} + \boldsymbol{\beta}
                    \end{aligned}
                $                                                               \\
                \cmidrule{1-1}
                \cmidrule{3-3}
                \makecell{\vspace{3ex}                                          \\ UnitNorm \\\vspace{3ex}}                                                                                   &                                                                              & $
                    \begin{aligned}
                        \tilde{\mathbf{X}} = & D^{\frac{k}{2}} \frac{\mathbf{X}}{\left\|\mathbf{X}\right\|} \\
                        \mathbf{Y} =         & \tilde{\mathbf{X}}
                    \end{aligned}
                $                                                               \\
                \bottomrule
            \end{tabular}
        \end{small}
    \end{center}
    \vskip -0.1in
\end{table}

\subsection{Feasibility of switching the order of normalization and projection in theoretical analysis} \label{sec:feasibility}

Let $\mathbf{X} \in \mathbb{R}^{L \times D}$ be a single sequence of token vectors, and the normalization operation is given in the following form:

\begin{equation}
    f: \mathbf{X} \mapsto \frac{\mathbf{X} - \boldsymbol{\mu}}{\boldsymbol{\sigma}} \equiv \mathbf{X}\mathbf{W} + \mathbf{b}
\end{equation}
where $\boldsymbol{\mu}$ and $\boldsymbol{\sigma}$ are the mean and standard deviation of the input vector $\mathbf{X}$, respectively, and $\mathbf{W} = \boldsymbol{\sigma}^{-1}$ and $\mathbf{b} = \boldsymbol{\mu}\boldsymbol{\sigma}^{-1}$. Depending on the normalization method, the mean and standard deviation can be computed over different dimensions.

The projection in the attention mechanism maps the input vectors to query, key and value vectors, and here we only consider the query and key vectors for this discussion, which are computed as follows:

\begin{equation}
    \begin{aligned}
        \mathbf{Q} = & \mathbf{X}\mathbf{W}_Q + \mathbf{b}_Q \\
        \mathbf{K} = & \mathbf{X}\mathbf{W}_K + \mathbf{b}_K
    \end{aligned}
\end{equation}
where $\mathbf{W}_Q, \mathbf{W}_K \in \mathbb{R}^{D \times D}$ are the projection matrices and $\mathbf{b}_Q, \mathbf{b}_K \in \mathbb{R}^{D}$ are the bias vectors for query and key, respectively. As the normalization and projection are both linear operations, we can combine them into a single linear operation as follows:

\begin{equation}
    \begin{aligned}
        \mathbf{Y} = & \tilde{\mathbf{X}}\mathbf{W}_Y + \mathbf{b}_Y                                                      \\
        =            & \left(\mathbf{X}\mathbf{W} + \mathbf{b}\right)\mathbf{W}_Y + \mathbf{b}_Y                          \\
        =            & \mathbf{X}\left(\mathbf{W}\mathbf{W}_Y\right) + \left(\mathbf{b}\mathbf{W}_Y + \mathbf{b}_Y\right)
    \end{aligned}
\end{equation}
for $Y \in \{Q, K\}$. Therefore, there must exist some $\mathbf{W}'$, $\mathbf{b}'$, $\mathbf{W}'_Y$ and $\mathbf{b}'_Y$ such that:

\begin{equation}
    \begin{aligned}
        \mathbf{W}'_Y\mathbf{W}' =              & \mathbf{W}\mathbf{W}_Y                \\
        \mathbf{b}_Y\mathbf{W}' + \mathbf{b}' = & \mathbf{b}\mathbf{W}_Y + \mathbf{b}_Y
    \end{aligned}
\end{equation}

Therefore, the order of normalization and projection does not affect the theoretical analysis. And in favor of simplicity, we can assume the normalization is performed after the projection.

%% file: 940.appendix.figure.tex
\section{Supplementary Figures} \label{sec:supplementary-figures}

\begin{figure}[htbp]
    \vskip 0.2in
    \begin{center}
        \centerline{\includegraphics[width=0.75\linewidth]{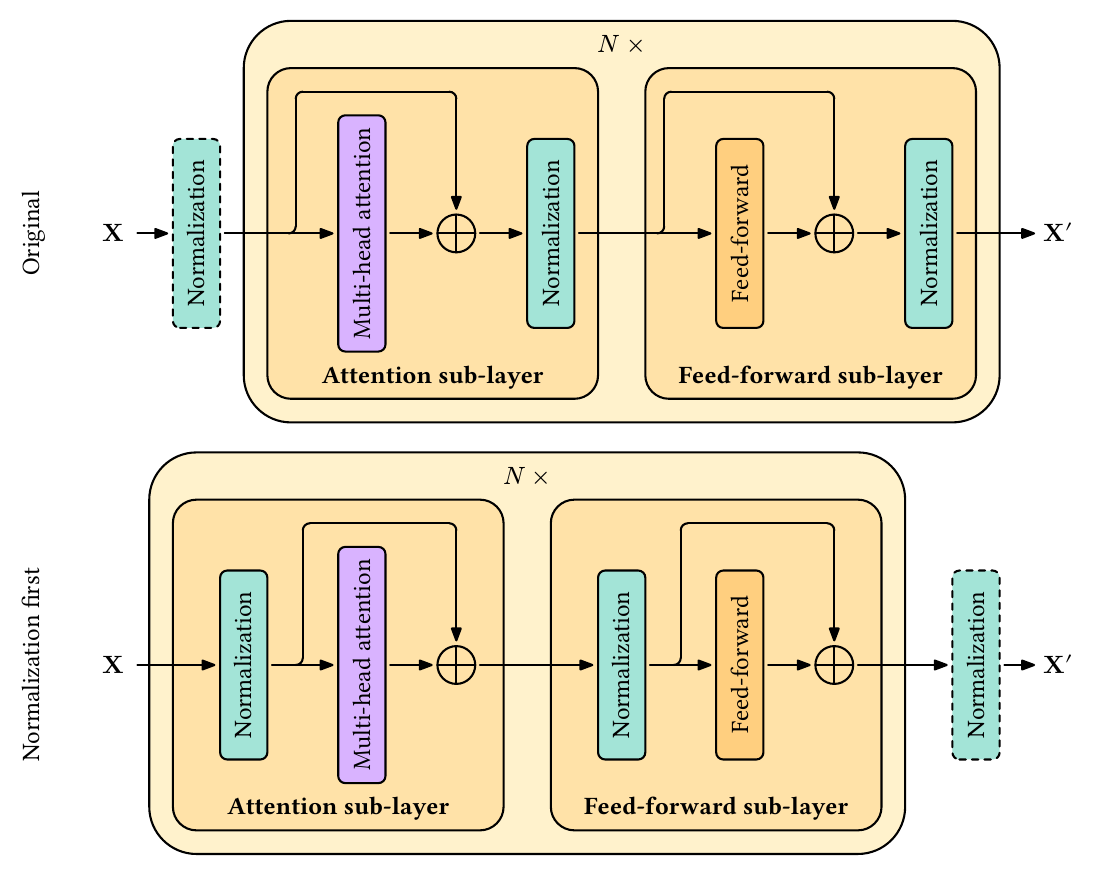}}
        \caption{Transformer layer architecture. The original architecture is equivalent to a normalization-first sub-layer design for simpler analysis.}
        \label{fig:trans-layer}
    \end{center}
    \vskip -0.2in
\end{figure}

\begin{figure}[htbp]
    \vskip 0.2in
    \begin{center}
        \centerline{\includegraphics[width=0.8\linewidth]{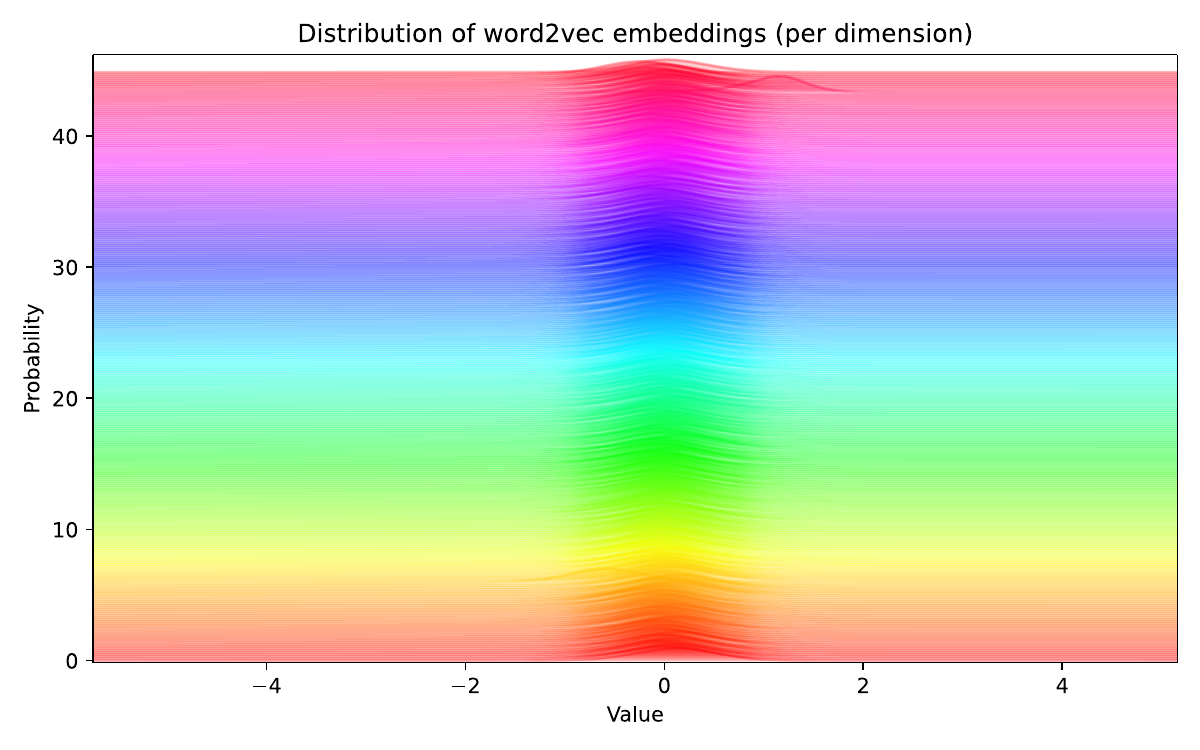}}
        \caption{Distribution of values in each dimension of the word2vec embedding. The word2vec embedding is a 300-dimensional vector, and the distribution follows a normal distribution with means mostly around 0.}
        \label{fig:word2vec-distribution}
    \end{center}
    \vskip -0.2in
\end{figure}

\begin{figure}[tbp]
    \vskip 0.2in
    \begin{center}
        \centerline{\includegraphics[width=0.9\linewidth]{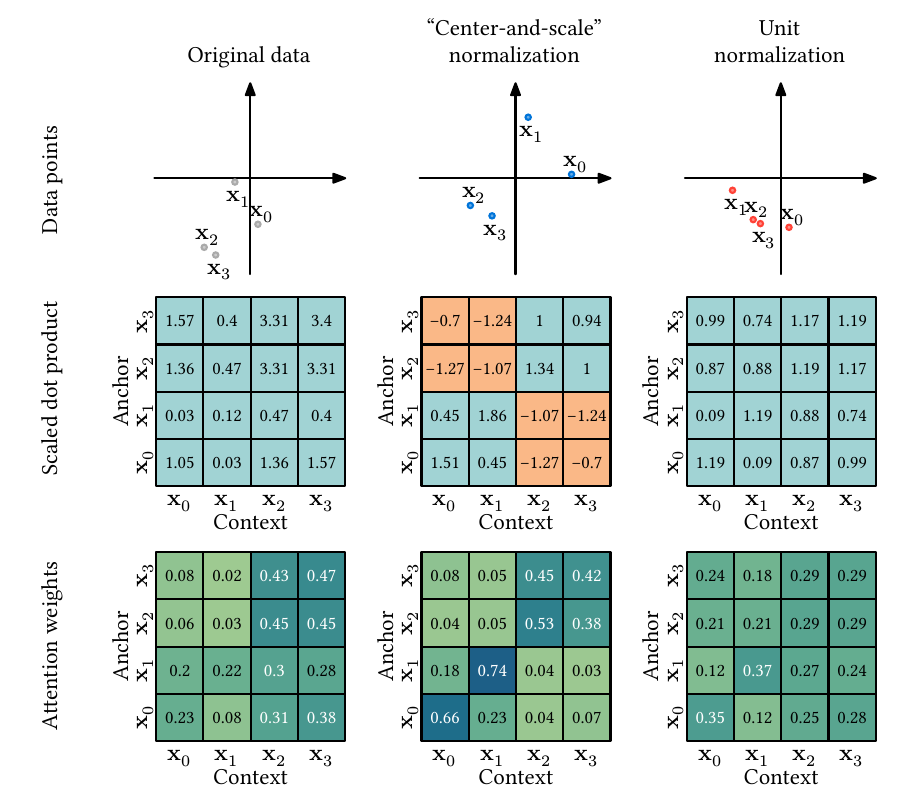}}
        \caption{Demonstration of the token shift and attention shift problems using artificial data. The $\mathbf{x}_0$ and $\mathbf{x}_1$ exhibit typical token shift as shifting away from their original quadrants, resulting in sign flip in scaled dot product (marked in orange), and leading to less attention weights distributed to $\mathbf{x}_2$ and $\mathbf{x}_3$ than original. Attention shift and sparse attention problem can also be observed as the maximum attention weight is altered from $\mathbf{x}_2$ and $\mathbf{x}_3$ to nearly solely onto themselves.}
        \label{fig:demo}
    \end{center}
    \vskip -0.3in
\end{figure}

\begin{figure}[htbp]
    \vskip 0.2in
    \begin{center}
        \centerline{\includegraphics[width=\linewidth]{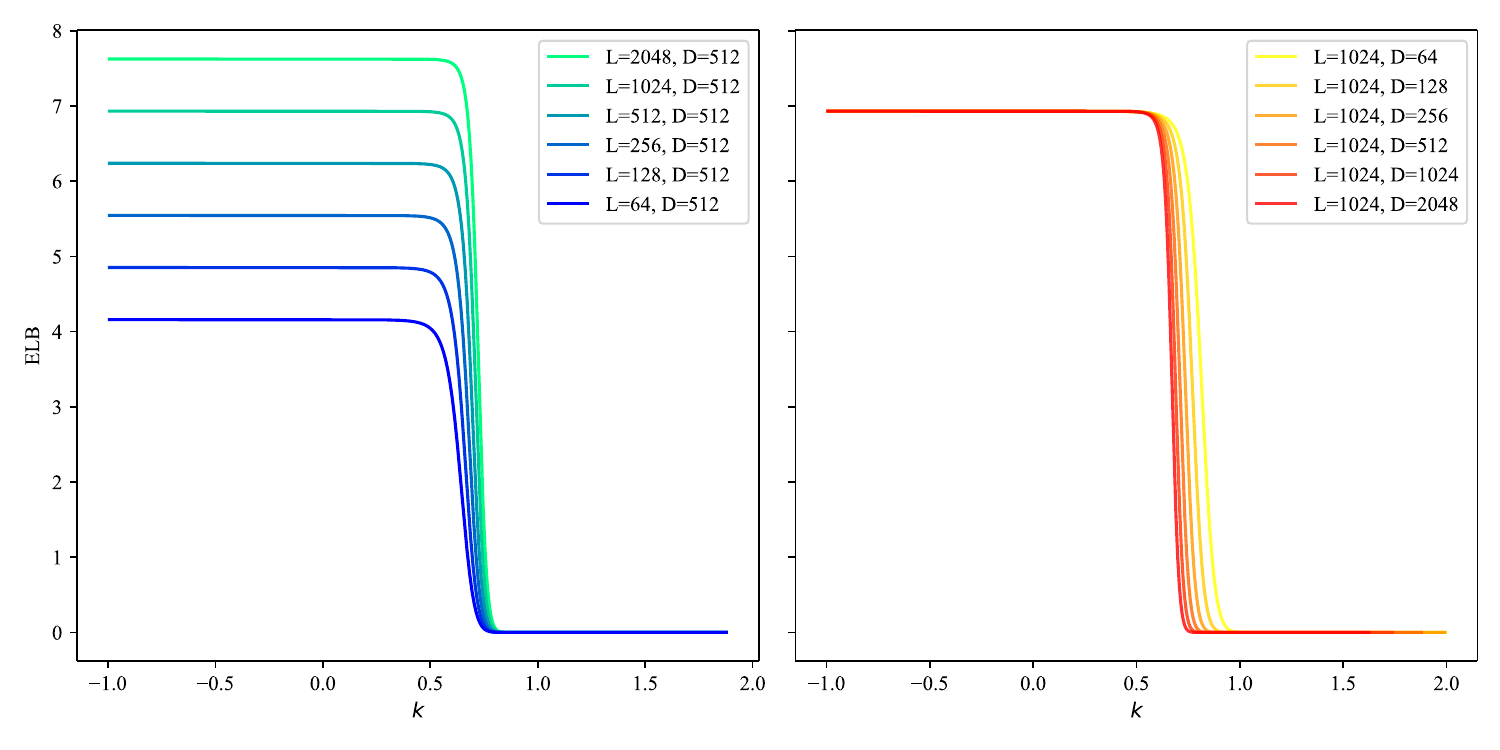}}
        \caption{Entropy lower bound (ELB) against $k$ for different $L, D$. The left figure shows the curve for fixed $D=512$ and varying $L$, and the right figure shows the curve for fixed $L=1024$ and varying $D$.}
        \label{fig:ELB-LD-curve}
    \end{center}
    \vskip -0.2in
\end{figure}

\begin{figure}[htbp]
    \vskip 0.2in
    \begin{center}
        \centerline{\includegraphics[width=0.6\linewidth]{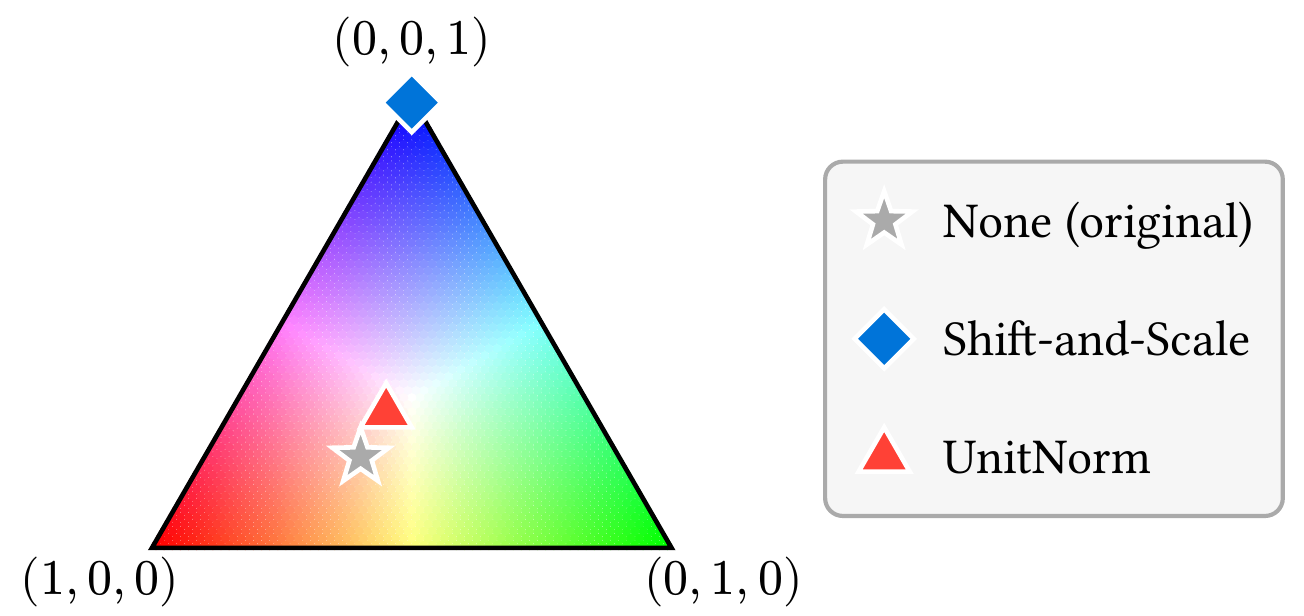}}
        \caption{Graphical representation of attention weights showing a simple scenario of 3 tokens. Each corner represents a one-hot distribution (red, blue and green) and the center representing a uniform distribution (white). Gray star, blue diamond, and red triangle mark the attention weights with no normalization, center-and-scale normalization, and UnitNorm, respectively.}
        \label{fig:attention-score-triangle}
    \end{center}
    \vskip -0.2in
\end{figure}

\begin{figure}[htbp]
    \vskip 0.2in
    \begin{center}
        \centerline{\includegraphics[width=\linewidth]{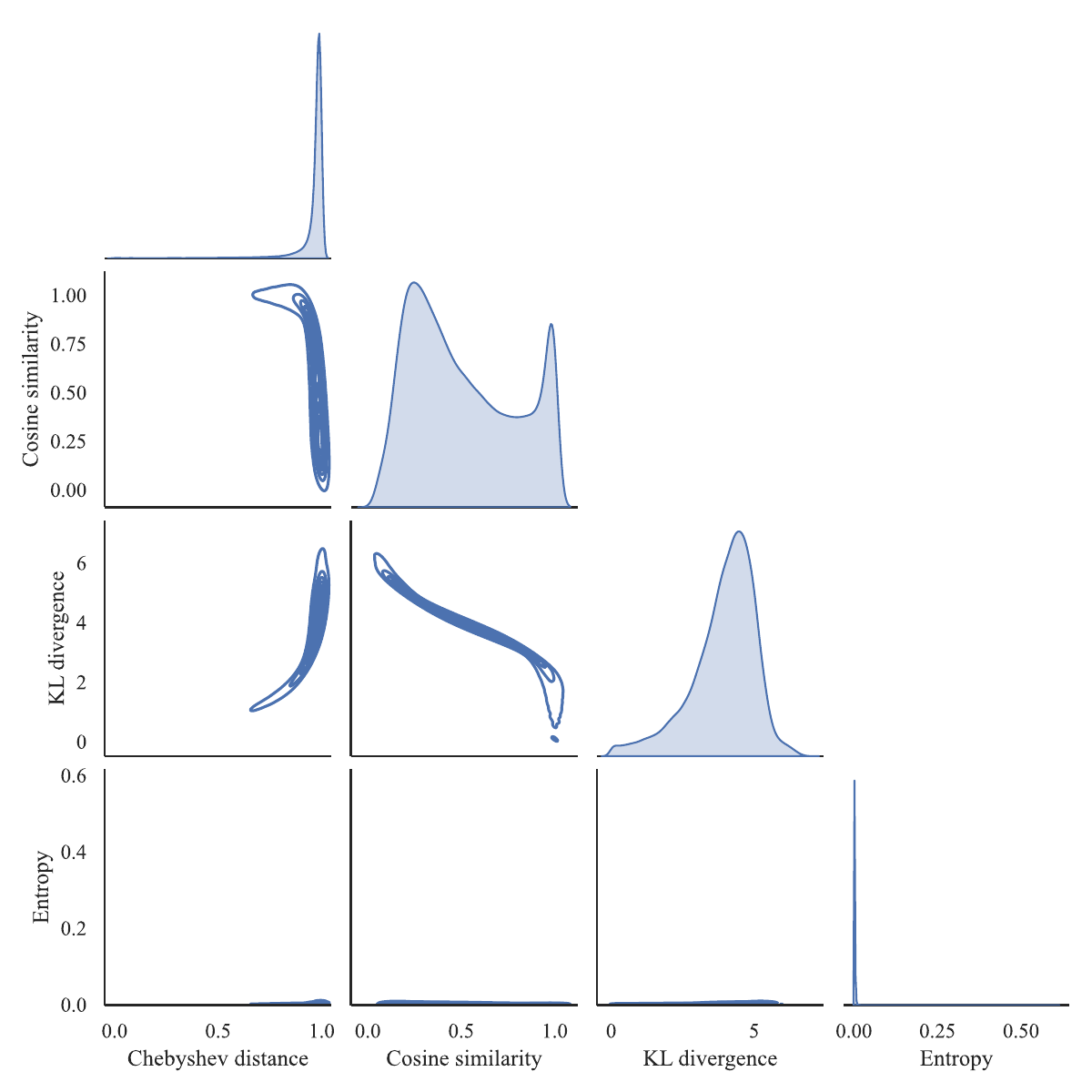}}
        \caption{Joint distribution of metrics for LayerNorm (practice). Metrics used are defined as in \cref{tab:metric-definition}.}
        \label{fig:word2vec-pair-layer-practice}
    \end{center}
    \vskip -0.2in
\end{figure}

\begin{figure}[htbp]
    \vskip 0.2in
    \begin{center}
        \centerline{\includegraphics[width=\linewidth]{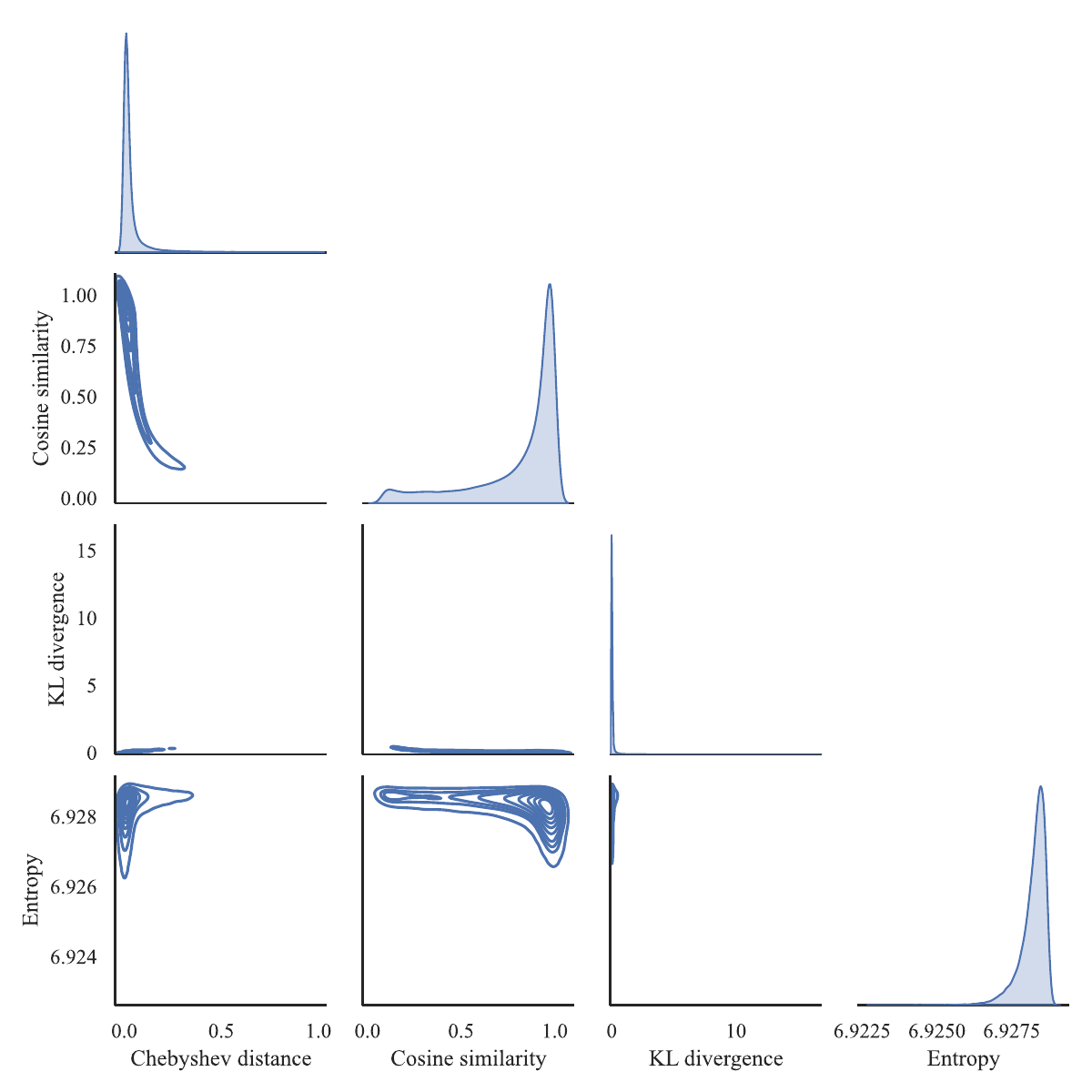}}
        \caption{Joint distribution of metrics for UnitNorm. Metrics used are defined as in \cref{tab:metric-definition}.}
        \label{fig:word2vec-pair-unit}
    \end{center}
    \vskip -0.2in
\end{figure}

\begin{figure}[htbp]
    \vskip 0.2in
    \begin{center}
        \centerline{\includegraphics[width=0.9\linewidth]{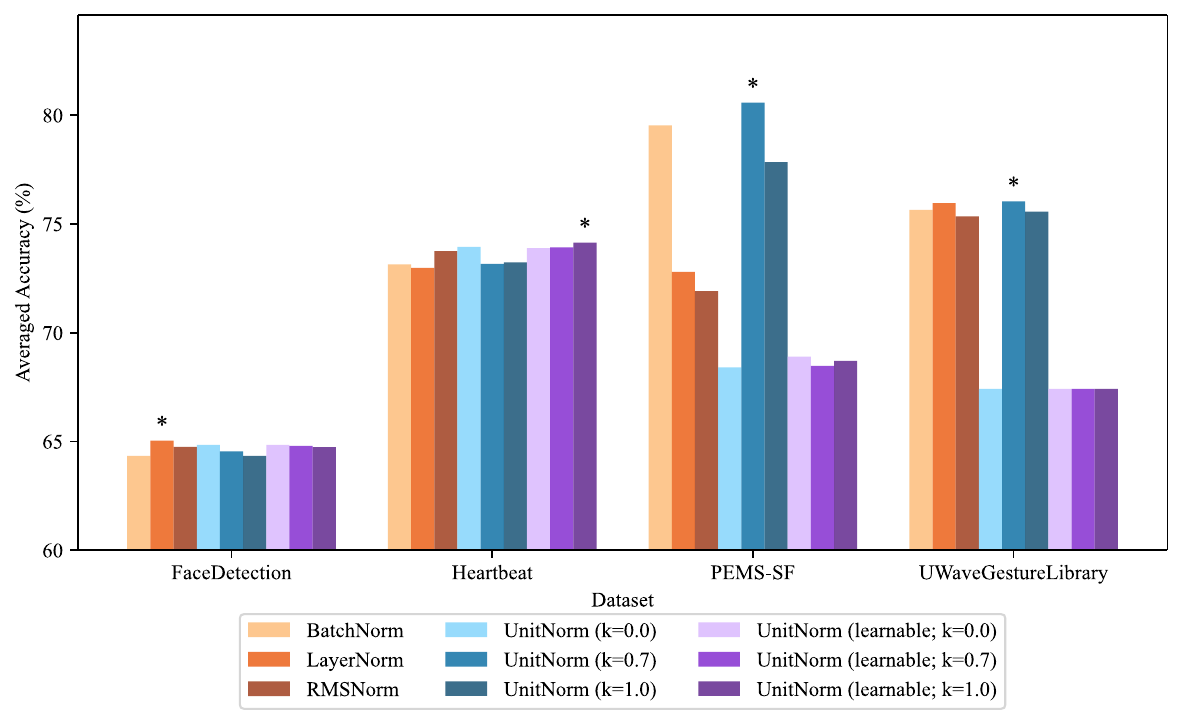}}
        \caption{Average rank of normalization methods on the classification tasks. X-axis: Dataset with different normalization, Y-axis: average rank over models. Ranks are computed based on the accuracy of each model on each task with different normalization methods (lower is better). $\ast$ indicates the best performing normalization method(s) on each task. UnitNorm and UnitNorm (learnable) outperform other normalization methods on 3 out of 5 datasets, showing its potential in classification tasks.}
        \label{fig:classification}
    \end{center}
    \vskip -0.2in
\end{figure}

\begin{figure}[htbp]
    \vskip 0.2in
    \begin{center}
        \centerline{\includegraphics[width=0.9\linewidth]{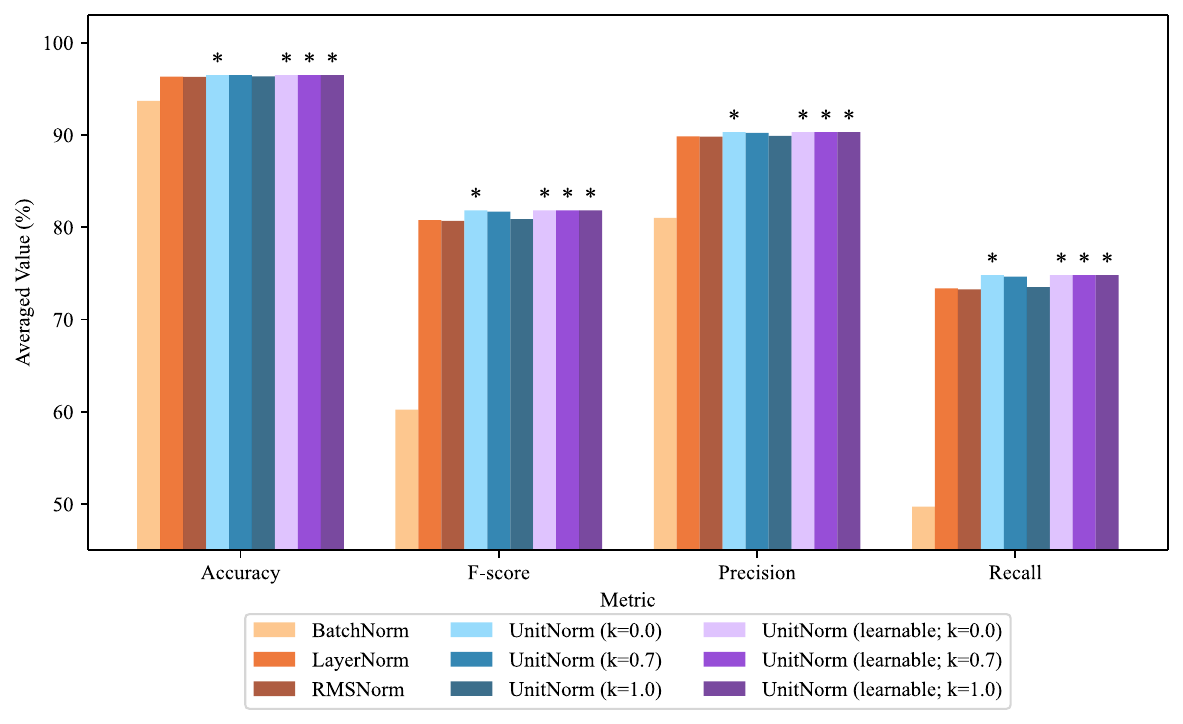}}
        \caption{Average rank of normalization methods on the anomaly detection tasks. X-axis: Metrics under different normalization, Y-axis: average rank over models. Ranks are computed based on every metric of each model with different normalization methods (lower is better). $\ast$ indicates the best performing normalization method(s) on each metric. UnitNorm and UnitNorm (learnable) show a dominating performance gain over the other normalization methods in all metrics.}
        \label{fig:anomaly-detection}
    \end{center}
    \vskip -0.2in
\end{figure}

\clearpage

%% file: 950.appendix.table.tex
\section{Supplementary Tables} \label{sec:supplementary-tables}

\begin{table}[!htb]
    \caption{Summary of long term forecasting benchmark settings. The sequence length is the number of historical time steps fed into the encoder, and the label length is the number of time steps fed into the decoder as the ground truth output of the decoder. The prediction length is the number of time steps to be predicted by the decoder.}
    \label{tab:long-term-forecast-dataset}
    \vskip 0.15in
    \begin{center}
        \begin{small}
\begin{tabular}{lccccccc}
                \toprule
                Datasets                                                 & \makecell{Feature                                                                                                                              \\number} & \makecell{Sequence\\length}      & \makecell{Label\\length}        & \makecell{Prediction length}                  & Metrics           & License        \\
                \midrule
                ETTh1, ETTh2 \cite{zhou_informer_2021}                   & 7                 & \multirow{3}{*}{384} & \multirow{3}{*}{96} & \multirow{3}{*}{96, 192, 384, 720} & \multirow{3}{*}{MSE, MAE} & CC BY-ND 4.0 \\
                ECL \cite{trindade_electricityloaddiagrams20112014_2015} & 321               &                      &                     &                                    &                           & CC BY 4.0    \\
                Exchange \cite{lai_modeling_2018}                        & 8                 &                      &                     &                                    &                           & N/A          \\
                \bottomrule
            \end{tabular}
        \end{small}
    \end{center}
    \vskip -0.1in
\end{table}

\begin{table}[!htb]
    \caption{Summary of classification benchmark settings. All datasets are from UEA Archive \cite{bagnall_uea_2018}. The sequence length is the number of time steps in each sequence fed into the encoder, and the prediction is made on the flattened output of the encoder by a fully connected layer.}
    \label{tab:classification-dataset}
    \vskip 0.15in
    \begin{center}
        \begin{small}
            \begin{tabular}{lccccc}
                \toprule
                Datasets            & Feature number & Class number & Sequence length     & Metrics                   & License              \\
                \midrule
                FaceDetection       & 144            & 5890         & \multirow{4}{*}{96} & \multirow{4}{*}{Accuracy} & \multirow{4}{*}{N/A} \\
                Heartbeat           & 61             & 204          &                     &                           &                      \\
                PEMS-SF             & 963            & 173          &                     &                           &                      \\
                UWaveGestureLibrary & 3              & 320          &                     &                           &                      \\
                \bottomrule
            \end{tabular}
        \end{small}
    \end{center}
    \vskip -0.1in
\end{table}

\begin{table}[!htb]
    \caption{Summary of anomaly detection benchmark settings. The sequence length is the number of time steps in each sequence fed into the model for reconstruction using MSE as loss. The threshold is determined by the distribution of reconstruction error on the training set, and the metrics are computed on the test set based on this threshold.}
    \label{tab:anomaly-detection-dataset}
    \vskip 0.15in
    \begin{center}
        \begin{small}
            \begin{tabular}{lccccc}
                \toprule
                Datasets                          & Feature number & Sequence length & Reconstruction error & Metrics                       & License \\
                \midrule
                MSL \cite{hundman_detecting_2018} & 55             & 100             & MSE                  & \makecell{Accuracy, F1-score,           \\Precision, Recall}  & N/A                      \\
                \bottomrule
            \end{tabular}
        \end{small}
    \end{center}
    \vskip -0.1in
\end{table}

\begin{table}[!htb]
    \caption{Summary of compute resources used for the experiments. Depending on the dataset and model, the GPU memory usage varies from 4G to 64G.}
    \label{tab:compute-resources}
    \vskip 0.15in
    \centering
    \begin{small}
        \begin{tabular}{ccccc}
            \toprule
            CPU                     & Memory & GPU                         & GPU Memory     \\
            \midrule
            AMD Threadripper 3995WX & 512G   & 4 $\times$ NVIDIA RTX A5000 & 4 $\times$ 24G \\
            \bottomrule
        \end{tabular}
    \end{small}
    \vskip -0.1in
\end{table}

\setlength{\tabcolsep}{2pt}

\begin{table}[!htb]
    \centering
    \caption{Long term forecasting test losses on different datasets using different models and normalization methods. For each dataset, prediction length, metric and for each model, the best performing normalization method(s) are bolded, and the second best are underlined.}
    \label{tab:long-term-forecast}
    \vskip 0.15in
    \scriptsize{
        \renewcommand{\arraystretch}{2.4}
        \input{tables/long-term-forcast}
    }
    \vskip -0.1in
\end{table}

\begin{table*}
    \centering
    \scriptsize{
        \renewcommand{\arraystretch}{2.4}
        \input{tables/long-term-forcast-2}
    }
\end{table*}

\begin{table}[!htb]
    \caption{Classification accuracies of different datasets using different models and normalization methods. For each dataset and for each model, the best performing normalization method(s) are bolded, and the second best are underlined.}
    \label{tab:classification}
    \vskip 0.15in
    \begin{center}
        \begin{small}
            \input{tables/classification}
        \end{small}
    \end{center}
    \vskip -0.1in
\end{table}

\begin{table}[!htb]
    \caption{Anomaly detection accuracies of MSL dataset using different models and normalization methods. For each metric and for each model, the best performing normalization method(s) are bolded, and the second best are underlined.}
    \label{tab:anomaly-detection}
    \vskip 0.15in
    \begin{center}
        \begin{small}
            \input{tables/anomaly-detection}
        \end{small}
    \end{center}
    \vskip -0.1in
\end{table}

\begin{table}[!htb]
    \caption{Metrics used for characterizing the distribution of attention scores from the original and normalized data, denoted as $\mathbf{A}_{n,i}$ and $\tilde{\mathbf{A}}_{n,i}$, respectively, for the $i$-th sample in the $n$-th batch.}
    \label{tab:metric-definition}
    \vskip 0.15in
    \begin{center}
        \begin{small}
            \begin{tabular}{clc}
                \toprule
                Metric             & Definition                                                                                                                                                                                               & Evaluation       \\
                \midrule
                Chebyshev distance & $D_{\text{Chebyshev}}\left(\mathbf{A}_{n,i},\tilde{\mathbf{A}}_{n,i}\right) = \max_{j=1}^{L}\left|\mathbf{A}_{n,i,j}-\tilde{\mathbf{A}}_{n,i,j}\right|$                                                  & Lower is Better  \\
                Cosine similarity  & $D_{\text{Cosine}}\left(\mathbf{A}_{n,i},\tilde{\mathbf{A}}_{n,i}\right) = \frac{\mathbf{A}_{n,i}^\top\tilde{\mathbf{A}}_{n,i}}{\left\|\mathbf{A}_{n,i}\right\|\left\|\tilde{\mathbf{A}}_{n,i}\right\|}$ & Higher is Better \\
                KL divergence      & $D_{\text{KL}}\left(\mathbf{A}_{n,i} \| \tilde{\mathbf{A}}_{n,i}\right) = \sum_{j=1}^{L}\mathbf{A}_{n,i,j}\left(\log\mathbf{A}_{n,i,j} - \log{\tilde{\mathbf{A}}_{n,i,j}}\right)$                        & Lower is Better  \\
                Entropy            & $E\left(\tilde{\mathbf{A}}_{n,i}\right) = -\sum_{j=1}^{L}\tilde{\mathbf{A}}_{n,i,j}\log\tilde{\mathbf{A}}_{n,i,j}$                                                                                       & Higher is Better \\
                \bottomrule
            \end{tabular}
        \end{small}
    \end{center}
    \vskip -0.1in
\end{table}

%% file: tables/long-term-forcast.tex
\begin{tabular}{cc|cccccccc|cccccccc|cccccccc|cccccccc}
\toprule
    & dataset & \multicolumn{8}{c}{ECL} & \multicolumn{8}{c}{ETTh1} & \multicolumn{8}{c}{ETTh2} & \multicolumn{8}{c}{Exchange} \\
    & prediction length & \multicolumn{2}{c}{96} & \multicolumn{2}{c}{192} & \multicolumn{2}{c}{336} & \multicolumn{2}{c}{720} & \multicolumn{2}{c}{96} & \multicolumn{2}{c}{192} & \multicolumn{2}{c}{336} & \multicolumn{2}{c}{720} & \multicolumn{2}{c}{96} & \multicolumn{2}{c}{192} & \multicolumn{2}{c}{336} & \multicolumn{2}{c}{720} & \multicolumn{2}{c}{96} & \multicolumn{2}{c}{192} & \multicolumn{2}{c}{336} & \multicolumn{2}{c}{720} \\
    & metric & \rotatebox[origin=c]{90}{MAE} & \rotatebox[origin=c]{90}{MSE} & \rotatebox[origin=c]{90}{MAE} & \rotatebox[origin=c]{90}{MSE} & \rotatebox[origin=c]{90}{MAE} & \rotatebox[origin=c]{90}{MSE} & \rotatebox[origin=c]{90}{MAE} & \rotatebox[origin=c]{90}{MSE} & \rotatebox[origin=c]{90}{MAE} & \rotatebox[origin=c]{90}{MSE} & \rotatebox[origin=c]{90}{MAE} & \rotatebox[origin=c]{90}{MSE} & \rotatebox[origin=c]{90}{MAE} & \rotatebox[origin=c]{90}{MSE} & \rotatebox[origin=c]{90}{MAE} & \rotatebox[origin=c]{90}{MSE} & \rotatebox[origin=c]{90}{MAE} & \rotatebox[origin=c]{90}{MSE} & \rotatebox[origin=c]{90}{MAE} & \rotatebox[origin=c]{90}{MSE} & \rotatebox[origin=c]{90}{MAE} & \rotatebox[origin=c]{90}{MSE} & \rotatebox[origin=c]{90}{MAE} & \rotatebox[origin=c]{90}{MSE} & \rotatebox[origin=c]{90}{MAE} & \rotatebox[origin=c]{90}{MSE} & \rotatebox[origin=c]{90}{MAE} & \rotatebox[origin=c]{90}{MSE} & \rotatebox[origin=c]{90}{MAE} & \rotatebox[origin=c]{90}{MSE} & \rotatebox[origin=c]{90}{MAE} & \rotatebox[origin=c]{90}{MSE} \\
\end{tabular}

%% file: tables/long-term-forcast-2.tex
\begin{tabular}{cc|cccccccc|cccccccc|cccccccc|cccccccc}
\toprule
    & dataset & \multicolumn{8}{c}{ECL} & \multicolumn{8}{c}{ETTh1} & \multicolumn{8}{c}{ETTh2} & \multicolumn{8}{c}{Exchange} \\
    & prediction length & \multicolumn{2}{c}{96} & \multicolumn{2}{c}{192} & \multicolumn{2}{c}{336} & \multicolumn{2}{c}{720} & \multicolumn{2}{c}{96} & \multicolumn{2}{c}{192} & \multicolumn{2}{c}{336} & \multicolumn{2}{c}{720} & \multicolumn{2}{c}{96} & \multicolumn{2}{c}{192} & \multicolumn{2}{c}{336} & \multicolumn{2}{c}{720} & \multicolumn{2}{c}{96} & \multicolumn{2}{c}{192} & \multicolumn{2}{c}{336} & \multicolumn{2}{c}{720} \\
    & metric & \rotatebox[origin=c]{90}{MAE} & \rotatebox[origin=c]{90}{MSE} & \rotatebox[origin=c]{90}{MAE} & \rotatebox[origin=c]{90}{MSE} & \rotatebox[origin=c]{90}{MAE} & \rotatebox[origin=c]{90}{MSE} & \rotatebox[origin=c]{90}{MAE} & \rotatebox[origin=c]{90}{MSE} & \rotatebox[origin=c]{90}{MAE} & \rotatebox[origin=c]{90}{MSE} & \rotatebox[origin=c]{90}{MAE} & \rotatebox[origin=c]{90}{MSE} & \rotatebox[origin=c]{90}{MAE} & \rotatebox[origin=c]{90}{MSE} & \rotatebox[origin=c]{90}{MAE} & \rotatebox[origin=c]{90}{MSE} & \rotatebox[origin=c]{90}{MAE} & \rotatebox[origin=c]{90}{MSE} & \rotatebox[origin=c]{90}{MAE} & \rotatebox[origin=c]{90}{MSE} & \rotatebox[origin=c]{90}{MAE} & \rotatebox[origin=c]{90}{MSE} & \rotatebox[origin=c]{90}{MAE} & \rotatebox[origin=c]{90}{MSE} & \rotatebox[origin=c]{90}{MAE} & \rotatebox[origin=c]{90}{MSE} & \rotatebox[origin=c]{90}{MAE} & \rotatebox[origin=c]{90}{MSE} & \rotatebox[origin=c]{90}{MAE} & \rotatebox[origin=c]{90}{MSE} & \rotatebox[origin=c]{90}{MAE} & \rotatebox[origin=c]{90}{MSE} \\
\bottomrule
\end{tabular}

%% file: tables/classification.tex
\begin{tabular}{cc|cccc}
\toprule
 & dataset & \makecell{Face\\Detection} & Heartbeat & PEMS-SF & \makecell{UWave\\Gesture\\Library} \\
\midrule
\multirow{5}{*}{\rotatebox[origin=c]{90}{BatchNorm}} & Crossformer & \textcolor{gray}{50.435} & \textbf{75.122} & \textbf{68.401} & \textbf{83.438} \\
 & FEDformer & \textcolor{gray}{68.275} & \underline{73.984} & \textcolor{gray}{78.035} & \textcolor{gray}{47.812} \\
 & Informer & \textcolor{gray}{68.606} & \textcolor{gray}{73.984} & \textbf{87.476} & \textcolor{gray}{82.083} \\
 & PatchTST & \textcolor{gray}{65.683} & \textcolor{gray}{66.992} & \textcolor{gray}{79.576} & \textcolor{gray}{81.354} \\
 & Transformer & \textcolor{gray}{68.663} & \underline{75.610} & \underline{84.200} & \textcolor{gray}{83.542} \\
\midrule
\multirow{5}{*}{\rotatebox[origin=c]{90}{LayerNorm}} & Crossformer & \textbf{52.176} & \textcolor{gray}{73.008} & \textcolor{gray}{26.397} & \underline{82.708} \\
 & FEDformer & \underline{68.861} & \textcolor{gray}{73.659} & \textcolor{gray}{84.393} & \underline{48.438} \\
\bottomrule
\end{tabular}

%% file: tables/anomaly-detection.tex
\begin{tabular}{cc|cccc}
\toprule
 & metric & Accuracy & F-score & Precision & Recall \\
\midrule
\multirow{5}{*}{\rotatebox[origin=c]{90}{BatchNorm}} & Crossformer & \textcolor{gray}{93.507} & \textcolor{gray}{59.110} & \textcolor{gray}{81.410} & \textcolor{gray}{47.903} \\
 & FEDformer & \textcolor{gray}{95.543} & \textcolor{gray}{75.820} & \textcolor{gray}{88.630} & \textcolor{gray}{66.240} \\
 & Informer & \textcolor{gray}{93.040} & \textcolor{gray}{56.927} & \textcolor{gray}{81.760} & \textcolor{gray}{43.733} \\
 & PatchTST & \textcolor{gray}{95.947} & \textcolor{gray}{78.613} & \textcolor{gray}{88.603} & \textcolor{gray}{70.650} \\
 & Transformer & \textcolor{gray}{90.417} & \textcolor{gray}{30.680} & \textcolor{gray}{64.623} & \textcolor{gray}{20.123} \\
\midrule
\multirow{5}{*}{\rotatebox[origin=c]{90}{LayerNorm}} & Crossformer & \textcolor{gray}{96.313} & \textcolor{gray}{80.640} & \textcolor{gray}{90.330} & \textcolor{gray}{72.823} \\
 & FEDformer & \textbf{96.603} & \textbf{82.427} & \textbf{90.697} & \textbf{75.537} \\
\bottomrule
\end{tabular}

%% file: 999.checklist.tex
\section*{NeurIPS Paper Checklist}

\begin{enumerate}

    \item {\bf Claims}
    \item[] Question: Do the main claims made in the abstract and introduction accurately reflect the paper's contributions and scope?
    \item[] Answer: \answerYes{} 
    \item[] Justification: We identify the critical issues of token shift, attention shift and re-evaluate the attention pattern problem in \cref{sec:challenges}, and provide experimental results in \cref{sec:experiments} to demonstrate the effectiveness of UnitNorm in addressing these challenges.
    \item[] Guidelines:
          \begin{itemize}
              \item The answer NA means that the abstract and introduction do not include the claims made in the paper.
              \item The abstract and/or introduction should clearly state the claims made, including the contributions made in the paper and important assumptions and limitations. A No or NA answer to this question will not be perceived well by the reviewers.
              \item The claims made should match theoretical and experimental results, and reflect how much the results can be expected to generalize to other settings.
              \item It is fine to include aspirational goals as motivation as long as it is clear that these goals are not attained by the paper.
          \end{itemize}

    \item {\bf Limitations}
    \item[] Question: Does the paper discuss the limitations of the work performed by the authors?
    \item[] Answer: \answerYes{} 
    \item[] Justification: We provide the limitations of this study in \cref{sec:limitations}.
    \item[] Guidelines:
          \begin{itemize}
              \item The answer NA means that the paper has no limitation while the answer No means that the paper has limitations, but those are not discussed in the paper.
              \item The authors are encouraged to create a separate "Limitations" section in their paper.
              \item The paper should point out any strong assumptions and how robust the results are to violations of these assumptions (e.g., independence assumptions, noiseless settings, model well-specification, asymptotic approximations only holding locally). The authors should reflect on how these assumptions might be violated in practice and what the implications would be.
              \item The authors should reflect on the scope of the claims made, e.g., if the approach was only tested on a few datasets or with a few runs. In general, empirical results often depend on implicit assumptions, which should be articulated.
              \item The authors should reflect on the factors that influence the performance of the approach. For example, a facial recognition algorithm may perform poorly when image resolution is low or images are taken in low lighting. Or a speech-to-text system might not be used reliably to provide closed captions for online lectures because it fails to handle technical jargon.
              \item The authors should discuss the computational efficiency of the proposed algorithms and how they scale with dataset size.
              \item If applicable, the authors should discuss possible limitations of their approach to address problems of privacy and fairness.
              \item While the authors might fear that complete honesty about limitations might be used by reviewers as grounds for rejection, a worse outcome might be that reviewers discover limitations that aren't acknowledged in the paper. The authors should use their best judgment and recognize that individual actions in favor of transparency play an important role in developing norms that preserve the integrity of the community. Reviewers will be specifically instructed to not penalize honesty concerning limitations.
          \end{itemize}

    \item {\bf Theory Assumptions and Proofs}
    \item[] Question: For each theoretical result, does the paper provide the full set of assumptions and a complete (and correct) proof?
    \item[] Answer: \answerYes{} 
    \item[] Justification: Assumptions and proofs are provided in \cref{sec:method}, \cref{sec:feasibility} and \cref{sec:proofs}
    \item[] Guidelines:
          \begin{itemize}
              \item The answer NA means that the paper does not include theoretical results.
              \item All the theorems, formulas, and proofs in the paper should be numbered and cross-referenced.
              \item All assumptions should be clearly stated or referenced in the statement of any theorems.
              \item The proofs can either appear in the main paper or the supplemental material, but if they appear in the supplemental material, the authors are encouraged to provide a short proof sketch to provide intuition.
              \item Inversely, any informal proof provided in the core of the paper should be complemented by formal proofs provided in appendix or supplemental material.
              \item Theorems and Lemmas that the proof relies upon should be properly referenced.
          \end{itemize}

    \item {\bf Experimental Result Reproducibility}
    \item[] Question: Does the paper fully disclose all the information needed to reproduce the main experimental results of the paper to the extent that it affects the main claims and/or conclusions of the paper (regardless of whether the code and data are provided or not)?
    \item[] Answer: \answerYes{} 
    \item[] Justification: The design of UnitNorm is fully disclosed in \cref{sec:method}, and related code and data are provided in \url{https://anonymous.4open.science/r/UnitNorm-5B84}.
    \item[] Guidelines:
          \begin{itemize}
              \item The answer NA means that the paper does not include experiments.
              \item If the paper includes experiments, a No answer to this question will not be perceived well by the reviewers: Making the paper reproducible is important, regardless of whether the code and data are provided or not.
              \item If the contribution is a dataset and/or model, the authors should describe the steps taken to make their results reproducible or verifiable.
              \item Depending on the contribution, reproducibility can be accomplished in various ways. For example, if the contribution is a novel architecture, describing the architecture fully might suffice, or if the contribution is a specific model and empirical evaluation, it may be necessary to either make it possible for others to replicate the model with the same dataset, or provide access to the model. In general. releasing code and data is often one good way to accomplish this, but reproducibility can also be provided via detailed instructions for how to replicate the results, access to a hosted model (e.g., in the case of a large language model), releasing of a model checkpoint, or other means that are appropriate to the research performed.
              \item While NeurIPS does not require releasing code, the conference does require all submissions to provide some reasonable avenue for reproducibility, which may depend on the nature of the contribution. For example
                    \begin{enumerate}
                        \item If the contribution is primarily a new algorithm, the paper should make it clear how to reproduce that algorithm.
                        \item If the contribution is primarily a new model architecture, the paper should describe the architecture clearly and fully.
                        \item If the contribution is a new model (e.g., a large language model), then there should either be a way to access this model for reproducing the results or a way to reproduce the model (e.g., with an open-source dataset or instructions for how to construct the dataset).
                        \item We recognize that reproducibility may be tricky in some cases, in which case authors are welcome to describe the particular way they provide for reproducibility. In the case of closed-source models, it may be that access to the model is limited in some way (e.g., to registered users), but it should be possible for other researchers to have some path to reproducing or verifying the results.
                    \end{enumerate}
          \end{itemize}

    \item {\bf Open access to data and code}
    \item[] Question: Does the paper provide open access to the data and code, with sufficient instructions to faithfully reproduce the main experimental results, as described in supplemental material?
    \item[] Answer: \answerYes{} 
    \item[] Justification: See code in \url{https://anonymous.4open.science/r/UnitNorm-5B84}.
    \item[] Guidelines:
          \begin{itemize}
              \item The answer NA means that paper does not include experiments requiring code.
              \item Please see the NeurIPS code and data submission guidelines (\url{https://nips.cc/public/guides/CodeSubmissionPolicy}) for more details.
              \item While we encourage the release of code and data, we understand that this might not be possible, so “No” is an acceptable answer. Papers cannot be rejected simply for not including code, unless this is central to the contribution (e.g., for a new open-source benchmark).
              \item The instructions should contain the exact command and environment needed to run to reproduce the results. See the NeurIPS code and data submission guidelines (\url{https://nips.cc/public/guides/CodeSubmissionPolicy}) for more details.
              \item The authors should provide instructions on data access and preparation, including how to access the raw data, preprocessed data, intermediate data, and generated data, etc.
              \item The authors should provide scripts to reproduce all experimental results for the new proposed method and baselines. If only a subset of experiments are reproducible, they should state which ones are omitted from the script and why.
              \item At submission time, to preserve anonymity, the authors should release anonymized versions (if applicable).
              \item Providing as much information as possible in supplemental material (appended to the paper) is recommended, but including URLs to data and code is permitted.
          \end{itemize}

    \item {\bf Experimental Setting/Details}
    \item[] Question: Does the paper specify all the training and test details (e.g., data splits, hyperparameters, how they were chosen, type of optimizer, etc.) necessary to understand the results?
    \item[] Answer: \answerYes{} 
    \item[] Justification: Important training details are provided in \cref{sec:supplementary-tables}. Others remain the same as in \cite{wu_timesnet_2023}.
    \item[] Guidelines:
          \begin{itemize}
              \item The answer NA means that the paper does not include experiments.
              \item The experimental setting should be presented in the core of the paper to a level of detail that is necessary to appreciate the results and make sense of them.
              \item The full details can be provided either with the code, in appendix, or as supplemental material.
          \end{itemize}

    \item {\bf Experiment Statistical Significance}
    \item[] Question: Does the paper report error bars suitably and correctly defined or other appropriate information about the statistical significance of the experiments?
    \item[] Answer: \answerNo{} 
    \item[] Justification: Error bars are not reported, as the mean value are calculated over different model architectures given the same hyperparameter and the same normalization method. Therefore, calculating the standard deviation is doable but not meaningful as it is not following a clear distribution.
    \item[] Guidelines:
          \begin{itemize}
              \item The answer NA means that the paper does not include experiments.
              \item The authors should answer "Yes" if the results are accompanied by error bars, confidence intervals, or statistical significance tests, at least for the experiments that support the main claims of the paper.
              \item The factors of variability that the error bars are capturing should be clearly stated (for example, train/test split, initialization, random drawing of some parameter, or overall run with given experimental conditions).
              \item The method for calculating the error bars should be explained (closed form formula, call to a library function, bootstrap, etc.)
              \item The assumptions made should be given (e.g., Normally distributed errors).
              \item It should be clear whether the error bar is the standard deviation or the standard error of the mean.
              \item It is OK to report 1-sigma error bars, but one should state it. The authors should preferably report a 2-sigma error bar than state that they have a 96\% CI, if the hypothesis of Normality of errors is not verified.
              \item For asymmetric distributions, the authors should be careful not to show in tables or figures symmetric error bars that would yield results that are out of range (e.g. negative error rates).
              \item If error bars are reported in tables or plots, The authors should explain in the text how they were calculated and reference the corresponding figures or tables in the text.
          \end{itemize}

    \item {\bf Experiments Compute Resources}
    \item[] Question: For each experiment, does the paper provide sufficient information on the computer resources (type of compute workers, memory, time of execution) needed to reproduce the experiments?
    \item[] Answer: \answerYes{} 
    \item[] Justification: The computation resources information is provided in \cref{tab:compute-resources}.
    \item[] Guidelines:
          \begin{itemize}
              \item The answer NA means that the paper does not include experiments.
              \item The paper should indicate the type of compute workers CPU or GPU, internal cluster, or cloud provider, including relevant memory and storage.
              \item The paper should provide the amount of compute required for each of the individual experimental runs as well as estimate the total compute.
              \item The paper should disclose whether the full research project required more compute than the experiments reported in the paper (e.g., preliminary or failed experiments that didn't make it into the paper).
          \end{itemize}

    \item {\bf Code Of Ethics}
    \item[] Question: Does the research conducted in the paper conform, in every respect, with the NeurIPS Code of Ethics \url{https://neurips.cc/public/EthicsGuidelines}?
    \item[] Answer: \answerYes{} 
    \item[] Justification: The authors follow the NeurIPS Code of Ethics in conducting the research.
    \item[] Guidelines:
          \begin{itemize}
              \item The answer NA means that the authors have not reviewed the NeurIPS Code of Ethics.
              \item If the authors answer No, they should explain the special circumstances that require a deviation from the Code of Ethics.
              \item The authors should make sure to preserve anonymity (e.g., if there is a special consideration due to laws or regulations in their jurisdiction).
          \end{itemize}

    \item {\bf Broader Impacts}
    \item[] Question: Does the paper discuss both potential positive societal impacts and negative societal impacts of the work performed?
    \item[] Answer: \answerNA{} 
    \item[] Justification: This paper is a theoretical study and does not have direct societal impacts.
    \item[] Guidelines:
          \begin{itemize}
              \item The answer NA means that there is no societal impact of the work performed.
              \item If the authors answer NA or No, they should explain why their work has no societal impact or why the paper does not address societal impact.
              \item Examples of negative societal impacts include potential malicious or unintended uses (e.g., disinformation, generating fake profiles, surveillance), fairness considerations (e.g., deployment of technologies that could make decisions that unfairly impact specific groups), privacy considerations, and security considerations.
              \item The conference expects that many papers will be foundational research and not tied to particular applications, let alone deployments. However, if there is a direct path to any negative applications, the authors should point it out. For example, it is legitimate to point out that an improvement in the quality of generative models could be used to generate deepfakes for disinformation. On the other hand, it is not needed to point out that a generic algorithm for optimizing neural networks could enable people to train models that generate Deepfakes faster.
              \item The authors should consider possible harms that could arise when the technology is being used as intended and functioning correctly, harms that could arise when the technology is being used as intended but gives incorrect results, and harms following from (intentional or unintentional) misuse of the technology.
              \item If there are negative societal impacts, the authors could also discuss possible mitigation strategies (e.g., gated release of models, providing defenses in addition to attacks, mechanisms for monitoring misuse, mechanisms to monitor how a system learns from feedback over time, improving the efficiency and accessibility of ML).
          \end{itemize}

    \item {\bf Safeguards}
    \item[] Question: Does the paper describe safeguards that have been put in place for responsible release of data or models that have a high risk for misuse (e.g., pretrained language models, image generators, or scraped datasets)?
    \item[] Answer: \answerNA{} 
    \item[] Justification: This paper focus on the theoretical side of the normalization method in time series Transformers and does not have high risks for misuse.
    \item[] Guidelines:
          \begin{itemize}
              \item The answer NA means that the paper poses no such risks.
              \item Released models that have a high risk for misuse or dual-use should be released with necessary safeguards to allow for controlled use of the model, for example by requiring that users adhere to usage guidelines or restrictions to access the model or implementing safety filters.
              \item Datasets that have been scraped from the Internet could pose safety risks. The authors should describe how they avoided releasing unsafe images.
              \item We recognize that providing effective safeguards is challenging, and many papers do not require this, but we encourage authors to take this into account and make a best faith effort.
          \end{itemize}

    \item {\bf Licenses for existing assets}
    \item[] Question: Are the creators or original owners of assets (e.g., code, data, models), used in the paper, properly credited and are the license and terms of use explicitly mentioned and properly respected?
    \item[] Answer: \answerYes{} 
    \item[] Justification: The authors properly credit the original owners of the assets and respect the license and terms of use, despite some assets used in this paper are missing the license information.
    \item[] Guidelines:
          \begin{itemize}
              \item The answer NA means that the paper does not use existing assets.
              \item The authors should cite the original paper that produced the code package or dataset.
              \item The authors should state which version of the asset is used and, if possible, include a URL.
              \item The name of the license (e.g., CC-BY 4.0) should be included for each asset.
              \item For scraped data from a particular source (e.g., website), the copyright and terms of service of that source should be provided.
              \item If assets are released, the license, copyright information, and terms of use in the package should be provided. For popular datasets, \url{paperswithcode.com/datasets} has curated licenses for some datasets. Their licensing guide can help determine the license of a dataset.
              \item For existing datasets that are re-packaged, both the original license and the license of the derived asset (if it has changed) should be provided.
              \item If this information is not available online, the authors are encouraged to reach out to the asset's creators.
          \end{itemize}

    \item {\bf New Assets}
    \item[] Question: Are new assets introduced in the paper well documented and is the documentation provided alongside the assets?
    \item[] Answer: \answerNA{} 
    \item[] Justification: This paper does not introduce new assets.
    \item[] Guidelines:
          \begin{itemize}
              \item The answer NA means that the paper does not release new assets.
              \item Researchers should communicate the details of the dataset/code/model as part of their submissions via structured templates. This includes details about training, license, limitations, etc.
              \item The paper should discuss whether and how consent was obtained from people whose asset is used.
              \item At submission time, remember to anonymize your assets (if applicable). You can either create an anonymized URL or include an anonymized zip file.
          \end{itemize}

    \item {\bf Crowdsourcing and Research with Human Subjects}
    \item[] Question: For crowdsourcing experiments and research with human subjects, does the paper include the full text of instructions given to participants and screenshots, if applicable, as well as details about compensation (if any)?
    \item[] Answer: \answerNA{} 
    \item[] Justification: This paper does not involve crowdsourcing nor research with human subjects.
    \item[] Guidelines:
          \begin{itemize}
              \item The answer NA means that the paper does not involve crowdsourcing nor research with human subjects.
              \item Including this information in the supplemental material is fine, but if the main contribution of the paper involves human subjects, then as much detail as possible should be included in the main paper.
              \item According to the NeurIPS Code of Ethics, workers involved in data collection, curation, or other labor should be paid at least the minimum wage in the country of the data collector.
          \end{itemize}

    \item {\bf Institutional Review Board (IRB) Approvals or Equivalent for Research with Human Subjects}
    \item[] Question: Does the paper describe potential risks incurred by study participants, whether such risks were disclosed to the subjects, and whether Institutional Review Board (IRB) approvals (or an equivalent approval/review based on the requirements of your country or institution) were obtained?
    \item[] Answer: \answerNA{} 
    \item[] Justification: This paper does not involve studies on human subjects.
    \item[] Guidelines:
          \begin{itemize}
              \item The answer NA means that the paper does not involve crowdsourcing nor research with human subjects.
              \item Depending on the country in which research is conducted, IRB approval (or equivalent) may be required for any human subjects research. If you obtained IRB approval, you should clearly state this in the paper.
              \item We recognize that the procedures for this may vary significantly between institutions and locations, and we expect authors to adhere to the NeurIPS Code of Ethics and the guidelines for their institution.
              \item For initial submissions, do not include any information that would break anonymity (if applicable), such as the institution conducting the review.
          \end{itemize}

\end{enumerate}